\documentclass[letterpaper]{article} 
\usepackage{aaai2026}  
\usepackage{times}  
\usepackage{helvet}  
\usepackage{courier}  
\usepackage[hyphens]{url}  
\usepackage{graphicx} 
\usepackage{natbib}  
\usepackage{caption} 
\usepackage{algorithm}
\usepackage{algorithmic}

\usepackage{newfloat}
\usepackage{listings}
\DeclareCaptionStyle{ruled}{labelfont=normalfont,labelsep=colon,strut=off} 
\floatstyle{ruled}
\newfloat{listing}{tb}{lst}{}
\floatname{listing}{Listing}
\usepackage{amsmath}
\usepackage{tabularx}
\usepackage{booktabs}
\usepackage{multirow}
\usepackage{rotating}
\usepackage{enumitem}
\usepackage{makecell}
\usepackage{bm}
\usepackage[table]{xcolor}
\definecolor{figure_gray}{HTML}{F3F3F3}

\usepackage{placeins}

\title{Scaling Laws for Conditional Emergence of Multilingual Image Captioning via Generalization from Translation}
\author{
    Julian Spravil\textsuperscript{\rm 1,3}\thanks{Corresponding author: julian.spravil\,@\,iais.fraunhofer.de},
    Sebastian Houben\textsuperscript{\rm 2,1},
    Sven Behnke\textsuperscript{\rm 3,4,5,1}
}
\affiliations{
    \textsuperscript{\rm 1}Fraunhofer IAIS, Germany\\
    \textsuperscript{\rm 2}Institute for Artificial Intelligence and Autonomous Systems, University of Applied Sciences Bonn-Rhein-Sieg, Germany\\
    \textsuperscript{\rm 3}Autonomous Intelligent Systems, Computer Science Institute VI, University of Bonn, Germany\\
    \textsuperscript{\rm 4}Lamarr Institute for Machine Learning and Artificial Intelligence, Germany\\
    \textsuperscript{\rm 5}Center for Robotics, University of Bonn, Germany\\
}

\begin{document}

\maketitle

\begin{abstract}
Cross-lingual, cross-task transfer is challenged by task-specific data scarcity, which becomes more severe as language support grows and is further amplified in vision-language models (VLMs).
We investigate multilingual generalization in encoder-decoder transformer VLMs to enable zero-shot image captioning in languages encountered only in the translation task.
In this setting, the encoder must learn to generate generalizable, task-aware latent vision representations to instruct the decoder via inserted cross-attention layers.
To analyze scaling behavior, we train Florence-2 based and Gemma-2 based models (0.4B to 11.2B parameters) on a synthetic dataset using varying compute budgets.
While all languages in the dataset have image-aligned translations, only a subset of them include image captions.
Notably, we show that captioning can emerge using a language prefix, even when this language only appears in the translation task.
We find that indirect learning of unseen task-language pairs adheres to scaling laws that are governed by the multilinguality of the model, model size, and seen training samples.
Finally, we demonstrate that the scaling laws extend to downstream tasks, achieving competitive performance through fine-tuning in multimodal machine translation (Multi30K, CoMMuTE), lexical disambiguation (CoMMuTE), and image captioning (Multi30K, XM3600, COCO Karpathy).
\end{abstract}


\section{Introduction}

\begin{figure}[!t]
    \centering
    \includegraphics[width=0.9\linewidth]{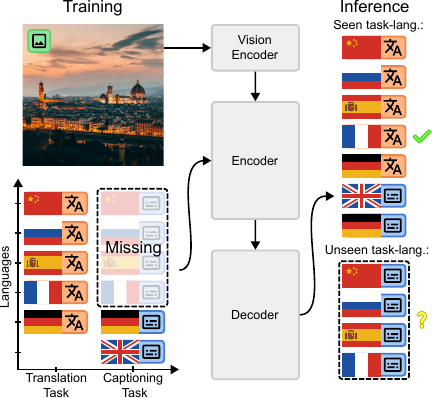}
    \caption{We train a vision-language model (VLM; middle) on an incomplete dataset (left) that covers the tasks image captioning (blue) and multimodal machine translation (orange). While En$\rightarrow$X translation is available for all languages, captioning data is limited to only English and German. The VLM generalizes to the missing captioning-language pairs with sufficient scale (right).}
    \label{fig:teaser}
\end{figure}

Multilingual image-to-text modeling is a fundamental step towards achieving universal accessibility of multimedia content.
Recent advancements in vision-language models (VLMs) demonstrate impressive results across various tasks such as image understanding and visual question answering~\cite{liu2023llava, xiao2024florence2, google2025gemma3}.
This progress is driven by the availability of large, primarily English vision-language datasets.
Two main approaches enable cross-lingual transfer, extending capabilities from English to other languages.
The first method involves fine-tuning multilingual models in a single language for a specific task, while keeping the embeddings and most layers of the language model frozen in order to retain its multilingual representations.~\cite{wu2019beto, chen2023mclip, futeral2025zerommt}.
Models such as mBERT~\cite{devlin2019bert} and NLLB~\cite{costajussa2022nllb} are common choices, representing established multilingual large language models (LLMs) and machine translation models (MTMs), respectively.
Evidence suggests that both multilingual and monolingual models learn language-agnostic representations, enabling cross-lingual transfer~\cite{libovicky2019how, souza2024measuring}.
The second method uses continuous pre-training on collected or generated multilingual data~\cite{gogoulou2022xlrmonolingual, qiu2022multilingual, futeral2024moscar}.

\begin{figure*}[t]
    \centering
    \includegraphics[trim={0 0.25cm 0 0.2cm},clip,width=0.9\linewidth]{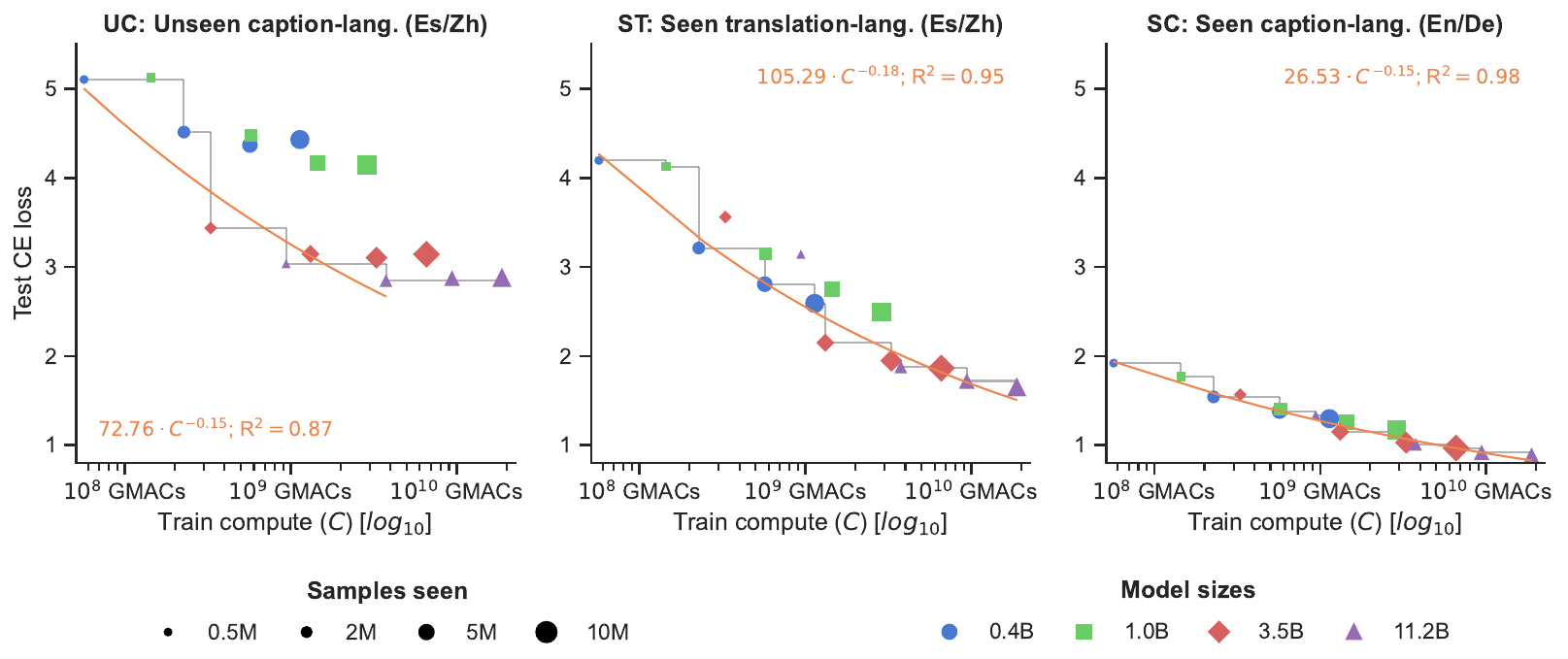}
    \caption{Test cross-entropy (CE) loss for various training compute budgets (GMACs, Giga multiply-accumulate operations). We show results for the test splits for unseen captioning (UC) in Spanish (Es) and Chinese (Zh), seen translation (ST) in the same languages, and seen captioning (SC) in English (En) and German (De). All models are trained for 0.5M, 2M, 5M, and 10M seen samples. Equation~\ref{eq:powerlaw1} is fitted to the points on the Pareto frontier (gray staircase graph). Higher compute budgets improve CE loss for UC (left), ST (middle), and SC (right). This suggests that translation facilitates generalization in captioning.}
    \label{fig:scaling_law}
\end{figure*}

Current methods are fundamentally constrained by their underlying models and data. 
By using pre-trained LLMs, they inherit the trade-off between performance and language coverage~\cite{conneau2020unsupervised}.
Generating data requires capable MTMs while collecting sufficient data is impractical.
Furthermore, multimodal approaches struggle to resolve lexical ambiguities (e.g., distinguishing between ``bat'' as an animal or sports equipment)~\cite{futeral2023commute}.
The dynamics of multilingual cross-task generalization, particularly at scale, remains largely unexplored, despite its implications for the necessity of collecting data for each task in every language.
To overcome these challenges, exploring systematic generalization~\cite{fodor1988connectionism,lake2023human} is a critical next step.

We explore the scaling laws of generalization within a realistic multimodal setting using a partially pre-trained encoder-decoder transformer~\cite{vaswani2017attention} and a standard training method.
Our goal is to learn a set of task and language combinations and transfer these capabilities to different task-language combinations in a zero-shot manner by scaling, as illustrated in Figure~\ref{fig:teaser}.
In summary, our main contributions are: 
\begin{itemize}
    \item We investigate the scaling laws of model performance on seen task-language data and its generalization to unseen task-language data, analyzing the effects of model size, number of seen training samples, and initial cross-entropy loss. We show that generalization by only learning translation to facilitate captioning is influenced not only by multilingual pre-training but also by model scale and seen training samples.
    \item We find that the observed scaling trends persist during fine-tuning, resulting in competitive performance across multiple benchmarks (Multi30K, CoMMuTE, COCO Karpathy, and XM3600).
    \item We present a modular encoder-decoder framework built on pre-trained VLMs and LLMs, with sizes ranging from 0.4B to 11.2B parameters.
    \item We propose a pipeline that generates synthetic multilingual image captions and aligns a text-only translation dataset to these images using contrastive VLMs and off-the-shelf MTMs.
\end{itemize}

\begin{figure*}[!t]
    \centering
    \includegraphics[trim={0 0.3cm 0 0.2cm},clip,width=0.9\linewidth]{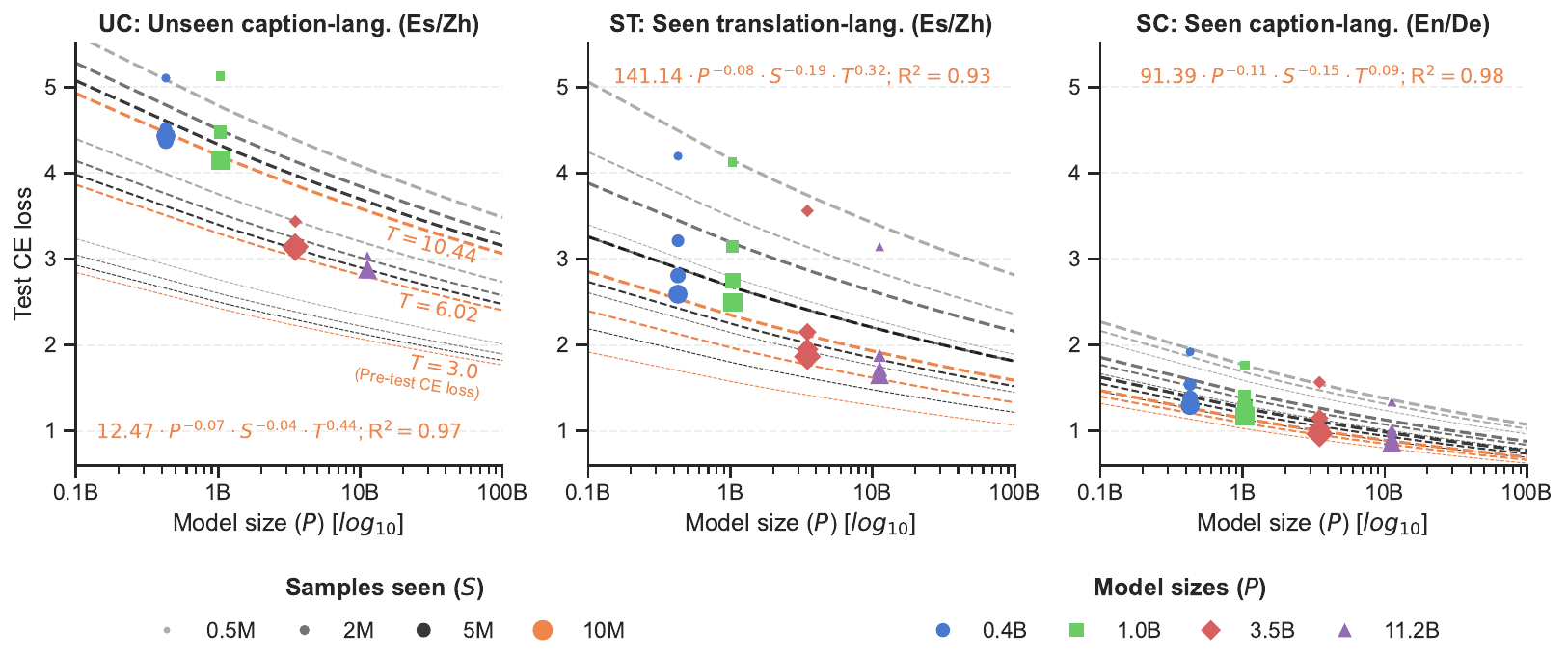}
    \caption{Test CE loss as a function of model size ($P$), number of seen samples ($S$), and initial CE loss ($T$) across the three test splits: UC, ST, and SC. The dashed lines represent the fitted functions for three values of $T$: $T = 10.44$ for Florence-2 based models, $T = 6.02$ for Gemma-2 based models, and $T = 3.0$ for a hypothetical highly multilingual VLM. Line thickness is proportional to the $T$ value. The measured results for all evaluated models are shown as points. The 10M seen-sample line is highlighted in orange, while lower sample counts are represented by progressively lighter shades of gray. Notably, for the UC task, test CE loss decreases as $P$ and $S$ increase and $T$ decreases.}
    \label{fig:power_law_prediction}
\end{figure*}

\section{Related Work}

\noindent
\textbf{Scaling laws.}
Following a power-law relationship, scaling laws enable predictable and efficient large-scale training and offer valuable insights into training dynamics.
These laws were first described for computer vision~\cite{sun2017revisiting} and have since been applied to natural language processing~\cite{kaplan2020scaling,ghorbani2021scaling,hoffmann2022chincilla,fernandes2023scaling}, transfer learning~\cite{hernandez2021scalingtransfer}, and contrastive vision-language learning~\cite{cherti2023reproducible}.
Current studies only focus on what is explicitly learned and not on what is implicitly learned.

\noindent
\textbf{Machine translation.}
The transformer~\cite{vaswani2017attention} has revolutionized machine translation, with impressive results through pre-training on extensive translation data~\cite{costajussa2022nllb}, non-parallel multilingual data~\cite{devlin2019bert}, and with weak supervision~\cite{conneau2020unsupervised}.
While models like NLLB~\cite{costajussa2022nllb} demonstrate significant success by supporting 200 languages, many translation directions remain under-resourced.
Several techniques have been used to address this issue, including language pivots~\cite{wu2007pivot}, the generation of pseudo labels~\cite{firat2016zerpresource}, and leveraging similar languages or parallel data~\cite{johnson2017google}.

\noindent
\textbf{Cross-lingual transfer.}
Multilingual LLMs~\cite{devlin2019bert, liu2020mbart} and MTMs~\cite{costajussa2022nllb} exhibit strong cross-lingual transfer performance, even when fine-tuned with monolingual data~\cite{wu2019beto, pires2019how, muennighoff2023crosslingual}.
Studies show that language-neutral and language-specific components develop and facilitate transfer~\cite{libovicky2019how,souza2024measuring}.

\noindent
\textbf{Multimodal multilingual learning.}
The issue of data scarcity is addressed by adapting multilingual models with machine translated data~\cite{futeral2025zerommt}, small multimodal multilingual datasets~\cite{mitzalis2021bertgen, futeral2023commute, hirasawa2023visual}, text-only translation data~\cite{mitzalis2021bertgen, hirasawa2023visual}, and image captioning as auxiliary task~\cite{mitzalis2021bertgen}.
Web-crawled multilingual vision data can unlock few-shot capabilities~\cite{futeral2024moscar}.
Adapting monolingual models can also be effective, although the size of the pre-training corpus can impact their final performance, while language similarity has little effect~\cite{gogoulou2022xlrmonolingual}.
Text-only models can solve some multimodal multilingual tasks that have few vision-dependent samples~\cite{hirasawa2023visual, futeral2023commute}.
Resolving lexical ambiguities requires additional context~\cite{futeral2023commute}.
Contrastive models~\cite{radford2021clip,carlsson2022mclip,chen2023mclip} excel at resolving ambiguities, whereas multilingual generative models struggle~\cite{futeral2023commute,futeral2025zerommt}.

\citet{muennighoff2023crosslingual} also perform cross-lingual transfer by fine-tuning LLMs without target-task data but rely on the model's existing multilingual capabilities.
In contrast, we investigate scaling laws for improving performance in a new task-language combination with respect to model size, seen training samples, and initial multilingual loss.

\section{Models and Datasets}

We construct partially pre-trained VLMs based on the pre-trained models Florence-2~\cite{xiao2024florence2} and Gemma-2~\cite{google2024gemma2}.
Florence-2 is an encoder-decoder VLM supporting tasks ranging from object detection to image captioning and is available with 0.2B and 0.8B parameters.
The encoder generates a sequence of tokens, representing both image and task, that is used to instruct the decoder via cross-attention layers.

To obtain larger model sizes, we combine Gemma-2, an LLM with sizes of 3B and 9B parameters, with the encoder of Florence-2.
The image-task encoder outputs are integrated into the decoder by inserted cross-attention layers that are weighted with a learnable parameter initialized with zero following Flamingo~\cite{alayrac2022flamingo}.

For the decoder, we reuse the tokenizer of Gemma-2 with a vocabulary size of 256k.
For the two smaller models with the Florence-2 decoder, we reinitialize the embedding layer and language modeling head to fit the Gemma-2 tokenizer using the method by \citet{gee2022fast}.
The encoder retains the original tokenizer and embeddings from Florence-2.

This leads to the standard transformer encoder-decoder VLM designed for continuous pre-training, available in sizes 0.4B, 1B, 3.5B, and 11.2B.

\begin{table}[bt]
\small
\centering
\begin{tabularx}{0.9\linewidth}{ccccc}
\toprule
\textbf{Task} & \textbf{Coefficient} & \textbf{Estimate} [95\% CI] & $\bm{p}$\textbf{-value} \\
\midrule
\multirow{3}{*}{SC} & $\beta_1$ & -0.59 [-0.79, -0.38] & $\bm{p} < 0.001$ \\
& $\beta_2$ & -0.72 [-0.80, -0.63] & $\bm{p} < 0.001$ \\
& $\beta_3$ & 0.10 [-0.10, 0.31] & $\bm{p} = 0.293$ \\
\midrule
\multirow{3}{*}{ST} & $\beta_1$ & -0.36 [-0.74, 0.03] & $\bm{p} = 0.068$ \\
& $\beta_2$ & -0.73 [-0.89, -0.57] & $\bm{p} < 0.001$ \\
& $\beta_3$ & 0.29 [-0.09, 0.68] & $\bm{p} = 0.125$ \\
\midrule
\multirow{3}{*}{UC} & $\beta_1$ & -0.41 [-0.69, -0.12] & $\bm{p} = 0.009$ \\
& $\beta_2$ & -0.23 [-0.35, -0.11] & $\bm{p} = 0.001$ \\
& $\beta_3$ & 0.57 [0.29, 0.85] & $\bm{p} < 0.001$ \\
\bottomrule
\end{tabularx}
\caption{Standardized coefficients for the second power law (Equation~\ref{eq:powerlaw2}) for UC, ST, and SC in log$_{10}$ space.}
\label{tab:beta_parameters}
\end{table}

\subsection{Continuous Pre-training Dataset}

We create a synthetic training dataset tailored for this study based on CC12M~\cite{changpinyo2019cc12m} and CCMatrix~\cite{schwenk2021ccmatrix}.
The dataset covers six languages: English (En), German (De), French (Fr), Spanish (Es), Russian (Ru), and Chinese (Zh).
CC12M contains 12M web-crawled images of which 10M are available.
We pair the images with generated image descriptions sourced from Hugging Face\footnote{\url{hf.co/datasets/CaptionEmporium/conceptual-captions-cc12m-llavanext}} and translated to German with NLLB-3.3B~\cite{costajussa2022nllb}.
CCMatrix is a web-crawled corpus covering 38 languages with 6.8B parallel sentences of which about 661M are aligned with English.
We extract translations from English to the aforementioned target languages.
Accelerated by Faiss~\cite{douze2024faiss}, we use CLIP-ViT-B/16~\cite{radford2021clip} to align the sentences with the images of CC12M via top-5 matching and subsequent deduplication.
See the appendix for more details.

In total, the training dataset contains 10M images aligned with 32M captions in En and De and 105M translation pairs for En$\rightarrow$\{De, Fr, Es, Ru, Zh\}.
To evaluate generalization, we intentionally omit captioning data for \{Fr, Es, Ru, Zh\}.

For the test set, we use a subset of 4.4K images from CC12M and create captioning data for two representative languages for the unseen task-language pairs (Es and Zh).
We divide the test set into three parts: unseen captioning (UC) with 4.4K Es and 3.5K Zh captions, seen translation (ST) with 4.1K En$\rightarrow$Es and 3.8K En$\rightarrow$Zh translations, and seen captioning (SC) with 4.4K En and 3.1K De captions.
Note that ``seen'' refers to the task-language combination being part of training, not the specific data instances.

\subsection{Downstream Tasks Dataset}
\label{sec:generalist_dataset}

We construct a fine-tuning dataset that includes a mix of downstream tasks with full language coverage, starting with the train split of Multi30K~\cite{elliott2016multi30k} for translation (Task\,1, En$\rightarrow$\{De, Fr\}).
For captioning, we include Multi30K (Task\,2) for short En and De captions, Image Paragraph~\cite{krause2017ip} for detailed captions, and DOCCI~\cite{onoe2024docci} for highly detailed descriptions.
Missing languages are added to the aforementioned datasets with neural machine translation~\cite{costajussa2022nllb}.
Additionally, we include the train/restval split of COCO Karpathy~\cite{chen2015cococap,karpathy2017deep}. 
In total, the fine-tuning dataset has 166K images with 1.6M captions of different styles and 145K translation samples covering all task-language combinations.

\begin{figure}[tb]
    \centering
    \includegraphics[trim={0 0.25cm 0 0.2cm},clip,width=0.87\linewidth]{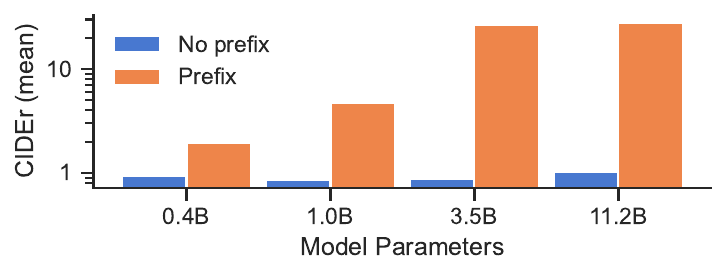}
    \caption{Effect of adding a prefix (Fr: ``La photo montre'', etc.) to the decoder input to unlock zero-shot captioning. Tested on the image captioning dataset XM3600 in the unseen languages Fr, Es, Ru, and Zh. The mean CIDEr over unseen languages significantly improves with the prefix.}
    \label{fig:zero_shot_xm3600_prefix}
\end{figure}

\begin{figure*}[t]
    \centering
    \includegraphics[trim={0 0.25cm 0 0.2cm},clip,width=0.95\linewidth]{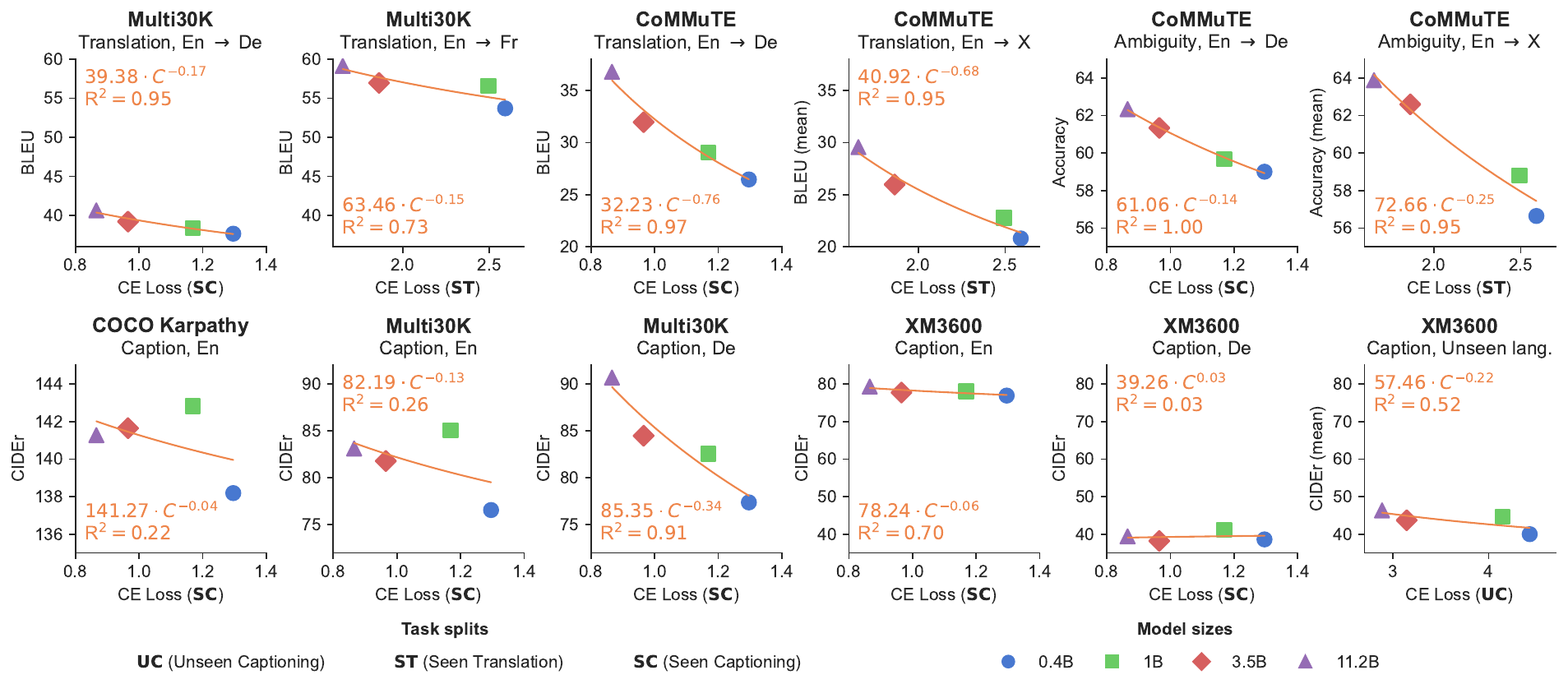}
    \caption{Downstream task performance with respect to CE loss, measured on the UC, ST, and SC tasks, depending on the type of downstream task. First row: Multi30K translation to De and Fr measured in BLEU (Task 1; mean over Test2016, Test2017 and AmbiguousCOCO splits), CoMMuTE translation and disambiguation for En$\rightarrow$De and En$\rightarrow$\{De, Fr, Ru, Zh\} measured in BLEU and accuracy, respectively. Second row: Captioning tasks measured with CIDEr: COCO Karpathy (En), Multi30K (En, De) (Task 2, Test2016), and XM3600 for En, De, and unseen languages (Fr, Es, Ru, Zh). We use a consistent y-axis scale for matching dataset and task.}
    \label{fig:scaling_law_downstream_task}
\end{figure*}

\begin{table*}[t]
\small
\centering
\begin{tabularx}{\linewidth}{l*{12}{>{\centering\arraybackslash}X}}
\toprule
& \multicolumn{2}{c}{\textbf{Multi30K}} & \multicolumn{2}{c}{\textbf{CoMMuTE}} & \multicolumn{2}{c}{\textbf{CoMMuTE}} & \textbf{COCO} & \multicolumn{2}{c}{\textbf{Multi30K}} & \multicolumn{3}{c}{\textbf{XM3600}} \\
& \multicolumn{2}{c}{\textbf{Translation}} & \multicolumn{2}{c}{\textbf{Translation}} & \multicolumn{2}{c}{\textbf{Ambiguity}} & \textbf{Caption} & \multicolumn{2}{c}{\textbf{Caption}} & \multicolumn{3}{c}{\textbf{Caption}} \\
\cmidrule(lr){2-3}
\cmidrule(lr){4-5}
\cmidrule(lr){6-7}
\cmidrule(lr){8-8}
\cmidrule(lr){9-10}
\cmidrule(lr){11-13}
& En$\rightarrow$De & En$\rightarrow$Fr & En$\rightarrow$De & En$\rightarrow$X & En$\rightarrow$De & En$\rightarrow$X & En & En & De & En & De & Unseen \\
\midrule
\textbf{Model} & BLEU & BLEU & BLEU & BLEU & Acc. & Acc. & CIDEr & CIDEr & CIDEr & CIDEr & CIDEr & CIDEr \\
\midrule
Gemma-3-12B* & 39.2 & 52.2 & \textbf{44.1} & \textbf{38.5} & \textbf{73.3} & \textbf{76.6} & 48.1 & 50.8 & 55.5 & 34.0 & 39.6 & 46.6 \\
Pixtral-12B* & 37.9 & 53.8 & 40.7 & 35.9 & 73.3 & 75.5 & 61.1 & 62.5 & 64.4 & 71.1 & 38.3 & \textbf{50.5} \\
\midrule
Baseline-6B* & 37.3 & 54.3 & 41.5 & 32.5 & 61.7 & 61.1 & \textbf{145.2} & 84.0 & 50.4 & \textbf{82.0} & 38.2 & 45.1 \\
\midrule
Florence-2-L &  & &  &  &  & & 143.3 &  &  & &  &   \\
PaliGemma-3B* &  &  &  &  &  & & 141.7 & \textbf{88.9} & 57.6 & 79.1 & 37.7 & 48.5 \\
\midrule
MOF & 24.9 & 35.1 &  &  & 63.7 & 66.5 &  &  &  &  &  &  \\
ZeroMMT-3.3B & 37.1 & 53.3 & & & 60.8 & 62.2 &  &  &  &  &  & \\
VGAMT & 37.4 & 58.4 &  &  & 57.1 &  &  &  &  &  &  &  \\
NLLB-3.3B* & 37.4 & 53.7 & 40.8 & 31.9 & 50.0 & 50.0 &  &  &  &  &  &  \\
\midrule
\rowcolor{figure_gray}
0.4B \textit{ft} (ours) & 37.7 & 53.7 & 26.5 & 20.8 & 59.0 & 56.6 & 138.2 & 76.6 & 77.4 & 76.9 & 38.6 & 40.0 \\
\rowcolor{figure_gray}
1.0B \textit{ft} (ours) & 38.4 & 56.6 & 29.0 & 22.8 & 59.7 & 58.8 & 142.8 & 85.1 & 82.6 & 78.0 & \textbf{41.2} & 44.7 \\
\rowcolor{figure_gray}
3.5B \textit{ft} (ours) & 39.2  & 56.9 & 32.0 & 26.0 & 61.3 & 62.6 & 141.6 & 81.8 & 84.5 & 77.7 & 38.2 & 43.7  \\ 
\rowcolor{figure_gray}
11.2B \textit{ft} (ours) & \textbf{40.7} & \textbf{59.1} & 36.8 & 29.6 & 62.3 & 63.9 & 141.3 & 83.1 & \textbf{90.7} & 79.3 & 39.4 & 46.3 \\
\midrule
\rowcolor{figure_gray} 
0.4B (ours) & 34.1 & 44.3 & 34.1 & 25.9 & 54.0 & 53.6 & 28.2 & 24.5 & 12.8 & $\text{24.8}^{\text{(31.3)}}$ & $\text{15.5}^{\text{(18.9)}}$ & $\text{0.8}^{\text{(1.7)}}$ \\
\rowcolor{figure_gray} 
1B (ours) & 35.3 & 47.4 & 35.4 & 27.1 & 54.7 & 54.1 & 21.3 & 17.9 & 9.4 & $\text{17.0}^{\text{(34.1)}}$ & $\text{13.1}^{\text{(20.0)}}$ & $\text{0.8}^{\text{(5.6)}}$ \\
\rowcolor{figure_gray} 
3.5B (ours) & 35.8 & 48.3 & 36.7 & 28.7 & 53.0 & 54.3 & 28.7 & 24.9 & 14.8 & $\text{24.4}^{\text{(38.2)}}$ & $\text{16.9}^{\text{(20.6)}}$ & $\text{0.8}^{\text{(24.1)}}$ \\
\rowcolor{figure_gray} 
11.2B (ours) & 36.6 & 50.9 & 39.5 & 29.8 & 52.7 & 53.5 & 30.5 & 26.1 & 15.6 & $\text{24.3}^{\text{(39.3)}}$ & $\text{17.6}^{\text{(20.1)}}$ & $\text{0.9}^{\text{(26.6)}}$ \\
\bottomrule
\end{tabularx}
\caption{Downstream task performance evaluated on Multi30K Task\,1 (translation, mean over the Test2016, Test2017, and AmbiguousCOCO splits), CoMMuTE translation and disambiguation (En$\rightarrow$\{De, Fr, Es, Ru, Zh\}), COCO Karpathy (En captioning), Multi30K Task\,2 (captioning), and XM3600 (captioning, Unseen contains \{Fr, Es, Ru, Zh\}). We report BLEU for translation, accuracy (Acc.) for disambiguation, CIDEr for captioning. \textbf{Bold} indicates best results, rows marked with * are evaluated by us, and values with a superscript number in braces are evaluated with a prefix.}
\label{tab:downstream_tasks}
\end{table*}

\section{Scaling Laws}
\label{sec:scaling_laws}

We explore the scaling laws of continuous pre-training in a multilingual multi-task scenario, where not all task-language combinations are given within the training data.
The relationship between cross-entropy (CE) loss and training compute can be described by a power law, where changes in model size and seen samples result in predictable, non-linear improvements~\cite{kaplan2020scaling,ghorbani2021scaling,hoffmann2022chincilla,fernandes2023scaling}.

Our first power law selects the Pareto frontier from all data points, an approach inspired by \citet{cherti2023reproducible}.
The scaling law to predict the CE loss $y$ from the training compute $C$ and error term $\epsilon$ is given by:
\begin{equation}
y = \alpha_0 C^{\alpha_1}  + \epsilon \text{,}
\label{eq:powerlaw1}
\end{equation}
where $\alpha_0$ and $\alpha_1$ are the parameters to be estimated.
The total computational cost $C$ in multiply–accumulate operations (MACs) is estimated by $C = S \cdot F \cdot \left( 1 + P_t / P \right)$, where $S$ is the number of seen training samples, $F$ is the forward pass MACs estimated with \texttt{fvcore}\footnote{\url{github.com/facebookresearch/fvcore}}, $P_t$ is the number of trainable parameters, and $P$ is the total number of parameters.

The second, multivariate power law predicts the CE loss $y$ based on the seen training samples $S$, the model parameters $P$, the mean initial CE loss $T$ of the base model on the UC and ST test sets, and the error term $\epsilon$:
\begin{equation}
y = \beta_0 P^{\beta_1} S^{\beta_2} T^{\beta_3} + \epsilon \text{,}
\label{eq:powerlaw2}
\end{equation}
where $\beta_0$, $\beta_1$, $\beta_2$, and $\beta_3$ are to be estimated.
In developing our model, we first considered a baseline using only variables $S$ and $P$ as commonly done in the literature.
However, the omission of variable $T$ led to a biased result.
We transform Equations \ref{eq:powerlaw1} and \ref{eq:powerlaw2} into $\log_{10}$ space to enable a linear regression analysis with ordinary least squares (OLS).

\noindent
\textbf{Training setup.}
We train all our models using AdamW~\cite{loshchilov2017adamw} with CE loss, a batch size of 1024, a weight decay of 0.01, and a learning rate of $1e^{-4}$, which is scheduled with a linear warm-up for 100 steps and cosine decay.
The input length for encoder and decoder is truncated to a maximum length of 128 and the image resolution is set to 224\,px.
Each model scale is trained for 500, 2K, 5K, and 10K iterations, where the latter corresponds to roughly one full epoch.
We use online balancing to sample equally from each task-language combination.
While the vision encoder is always frozen, we freeze the decoder layers of the 3.5B and 11.2B models as well. 
This means that the vision-task encoder, the inserted cross-attention layers, the language modeling head, and the embedding layer are trainable.
Using the \texttt{transformers} library~\cite{wolf-etal-2020-transformers}, we trained on a HPC node equipped with 4 NVIDIA H100 GPUs. 

\noindent
\textbf{Evaluation setup.}
We calculate the CE loss $y$ over the three test sets of our continuous pre-training dataset covering the settings: unseen captioning (UC) in Es and Zh, seen translation (ST) in Es and Zh, and seen captioning (SC) in En and De.
The initial CE loss $T$ is calculated as a measure of multilinguality on the UC and ST test sets before training is conducted on the untrained but restructured models.

\subsection{Results}

The CE loss of the runs with compute budget and the fit of the first power law (Equation~\ref{eq:powerlaw1}) are visualized in Figure~\ref{fig:scaling_law}.
The analysis confirms a strong fit for SC ($R^2{=}0.98$) and ST ($R^2{=}0.95$) and a slightly weaker fit for UC ($R^2{=}0.87$).
All models exhibit a clear inverse correlation between CE loss and train compute, which, while expected for seen tasks (SC and ST), suggests that models can generalize to unseen tasks (UC) in a language encountered only through translation.

Figure~\ref{fig:power_law_prediction} presents the observed unstandardized coefficients of the second power law (Equation~\ref{eq:powerlaw2}) along with extrapolations for larger and more multilingual models.
To compare the relative importance of predictors across models, we additionally report standardized coefficients (standardized across the combined SC, ST, and UC tasks) in Table~\ref{tab:beta_parameters}.
All three models have a strong fit with $R^2{=}0.98$, $R^2{=}0.93$, and $R^2{=}0.97$ for SC, ST, and UC, respectively.
Regression diagnostics indicate that OLS assumptions are satisfied.

The SC model is the standard setting with En and De captioning in the training and evaluation data.
The standardized coefficients reveal that both training samples $S$ ($\beta_2{=}-0.72$) and model size $P$ ($\beta_1{=}-0.59$) are negatively correlated with the test CE loss.
In contrast, a lower initial CE loss $T$ is weakly connected with a lower predicted loss ($\beta_3{=}0.10$), though this effect has high uncertainty.
Though the negative dependencies on $S$ and $P$ are consistent with standard scaling laws, our findings contradict those of \citet{zheng2024breaking}, who suggest that model size is more important than dataset size for the continuous pre-training of LLMs for cross-lingual transfer.
We found no evidence that either predictor is more important, as the 95\% confidence interval (CI) of the estimated difference between coefficients includes zero ($\beta_1{-}\beta_2{=}0.13$, 95\% CI [-0.09, 0.35]).
This suggests that known scaling behaviors may differ for multimodal tasks.

The ST model covers multilingual machine translation from En to Es and Zh.
In this setting, only the number of training samples $S$ ($\beta_2{=}-0.73$) has a negative effect on test CE loss.
The standardized coefficients for both model size $P$ ($\beta_1{=}-0.36$) and initial CE loss $T$ ($\beta_3{=}0.29$) remain inaccurate.
The lack of a clear effect is unexpected, as larger, more multilingual models are presumed to perform better. 
An explanation for this finding and the model's lower $R^2$ value is that machine translation may require separate terms for encoder and decoder parameters~\cite{ghorbani2021scaling}.
In our setup, the encoder is relatively small, which may introduce a bottleneck.

The UC model tests generalization to unseen image captioning in Es and Zh.
Standardized coefficients show a positive association for initial CE loss $T$ ($\beta_3{=}0.57$) and negative associations for model size $P$ ($\beta_1{=}-0.41$) and seen training samples $S$ ($\beta_2{=}-0.23$).
By comparing the magnitudes, we identify that the influence of initial CE loss $T$ is greater than that of training samples $S$ ($\beta_3{+}\beta_2{=}0.34$, 95\% CI [0.03, 0.65]).
In contrast, the differences in magnitude between $T$ and model size $P$ ($\beta_3{+}\beta_1{=}0.16$, 95\% CI [-0.39, 0.72]) and $P$ and seen training samples $S$ ($\beta_1{-}\beta_2{=}-0.18$, 95\% CI [-0.48, 0.13]) are not distinctive.
While larger models are known to perform well on zero-shot tasks, this is often attributed to potential dataset contamination~\cite{radford2019language}.
Our findings suggest that generalization is not merely an artifact of pre-training data contamination. 
Instead, overall model capacity and the quantity of observed, problem-related training data also play a critical role.

During inference, even though the captioning CE loss for unseen languages decreases, the models consistently fail to produce text in the intended target language.
Instead, they default to En or De, the two languages encountered during training for captioning.
We found that adding a small prefix to the decoder seeds the output of the model.
The effect is visualized in Figure~\ref{fig:zero_shot_xm3600_prefix} and shows the generation of captions without prior exposure to captioning data in those languages.
Qualitative examples can be found in the appendix.

We extrapolate the second power law to estimate the CE loss values for a larger, highly multilingual model with $P{=}\text{30B}$, $T{=}3.0$, and a fixed compute budget of $S{=}\text{10M}$.
We predict that this model could achieve a CE loss of $1.92$ with a 95\% prediction interval (PI) [$1.65$, $2.23$], $1.18$ with a 95\% PI [$0.89$, $1.57$], and $0.71$ with a 95\% PI [$0.63$, $0.80$] on UC, ST, and SC, respectively.

\noindent
\textbf{Key findings.} 
The insights of this scaling law study can be summarized as follows: CE loss is predicted by initial multilinguality $T$, model size $P$, and seen training samples $S$. For captioning with full coverage (SC), $P$ and $S$ contribute comparably; for translation with only translation supervision (ST), $S$ dominates; and for captioning with only translation supervision (UC), all three matter. Our results indicate that scaling reduces, but does not eliminate, the need for task-language supervision.

\noindent
\textbf{Limitations.} 
This study has several limitations.
First, our scaling-law analysis is based on only 16 experimental configurations, limiting the predictive capability of our findings.
Second, the derived power laws are specific to our experimental setup.
Additionally, other factors could influence the parameters, including: the number of languages that the model has to learn, the extensiveness of the pre-training, the synthetic nature of our data and potential style and domain gaps, the difficulty of tasks, and the effect of multiple tasks.

\section{Downstream Tasks}
\label{sec:downstream_tasks}

To evaluate if the scaling laws transfer, we train on a mix of downstream tasks designed to enhance multilingual translation and captioning.

\noindent
\textbf{Training setup.}
Starting from the models pre-trained for 10K steps, we train for an additional 5K steps using a similar setting. 
The image resolution is set to 768\,px, the batch size to 256, and the learning rate to $5e^{-5}$.

\noindent
\textbf{Evaluation setup.}
We evaluate our models on three tasks: image captioning, multimodal machine translation, and lexical disambiguation.
Image captioning is assessed on COCO Karpathy~\cite{chen2015cococap,karpathy2017deep} (5K images with five En captions), Multi30K (Task\,2, 1K images with five En and five De captions)~\cite{elliott2016multi30k}, and XM3600~\cite{thapliyal2022xm3600} (3.6K images, captions in 36 languages) evaluated with CIDEr~\cite{vedantam2015cider} using the \texttt{pycocoevalcap} toolkit\footnote{\url{https://github.com/salaniz/pycocoevalcap}}.
Following \citet{futeral2024moscar}, we apply segmentation with \texttt{stanza}~\cite{qi2020stanza} for languages without word boundaries.
Multimodal machine translation is assessed with BLEU~\cite{papineni2002bleu} (via SacreBLEU~\cite{post2019sacrebleu}) on the Multi30K (Task 1, Test2016, Test2017, AmbiguousCOCO splits)~\cite{elliott2016multi30k,elliott2017multi30ktask1test2017,barrault2018multi30ktest2018} and on CoMMuTE~\cite{futeral2023commute,futeral2025zerommt} (310 translations with images for context).
Finally, lexical disambiguation is assessed with accuracy on CoMMuTE.

\subsection{Results}

We perform OLS regressions in log$_{10}$ space to model downstream task performance as a function of the CE loss on our UC, ST, and SC test splits.
The resulting trend curves are plotted in Figure~\ref{fig:scaling_law_downstream_task}.
We observe a strong to moderate fit for most downstream tasks indicating that a lower CE loss on the UC, ST, and SC test splits generally translates to better downstream task performance.
For En tasks and De tasks with full task-language coverage, we observe weaker fits.
This is likely because performance has begun to plateau, approaching or even exceeding state-of-the-art results on XM3600 and Multi30K De captioning.
The captioning task on unseen languages in the Multi30K dataset shows a moderate fit ($R^2{=}0.52$).
This indicates that the CE loss on the UC task is likely a decent predictor for downstream task performance. 
However, these interpretations should be taken with caution due to the limited number of measurements.

The downstream task performance is detailed in Table~\ref{tab:downstream_tasks}. 
More detailed results can be found in the supplementary material.
To put our experimental results into perspective, we compare to a combination of BLIP-2~\cite{li2023blip2} and NLLB-3.3B~\cite{costajussa2022nllb} with context-enhanced translation, referred to as Baseline-6B.
For captioning, we include PaliGemma-3B~\cite{beyer2024paligemma} and Florence-2-L~\cite{xiao2024florence2}.
For translation, we use NLLB-3.3B, its multimodal extension ZeroMMT-3.3B~\cite{futeral2025zerommt}, Multilingual Open Flamingo (MOF)~\cite{futeral2024moscar}, and multilingual VGAMT~\cite{futeral2023commute}.
Furthermore, we include three state-of-the-art multilingual VLMs as references: Pixtral-12B~\cite{agrawal2024pixtral} and Gemma-3-12B~\cite{google2025gemma3}.
Note that baseline models vary in their degree of exposure to the downstream training data.

Before fine-tuning, the performance across all benchmarks of our models is relatively weak but improving with scale.
Translation performance is slightly worse than our baseline. 
However, the captioning metrics appear artificially low, likely due to a style and domain mismatch.
A language-specific prefix (see Figure~\ref{fig:zero_shot_xm3600_prefix}) resolves the complete failure on unseen-language captioning tasks (0.9 CIDEr without and 26.6 CIDEr with prefix for the 11.2B model on XM3600 unseen) while also boosting CIDEr scores for En and De.
The ``step down'' in captioning performance between 1.0B and 3.5B, along with the slight drop in performance of 1.0B in translation benchmarks, suggests that learning a multilingual embedding layer is more difficult than learning multimodal alignment.
The gap does not transfer to fine-tuned models, however.

Unsurprisingly, fine-tuning on the combined downstream task dataset leads to substantial scalable performance improvements across nearly all benchmarks.
Our model achieves the best performance for Multi30K Translation (40.7 and 59.1 BLEU for En$\rightarrow$\{De, Fr\}, respectively), and image captioning in De on Multi30K (90.7 CIDEr) and XM3600 (41.2 CIDEr), outperforming both specialized and more-capable foundation models.
On unseen languages in XM3600, our models are competitive, even surpassing Baseline-6B, and perform only slightly worse than Gemma-3-12B, despite not relying on extensive multilingual multimodal pre-training.
The overall best performance is achieved by Pixtral-12B (50.5 CIDEr).

The BLEU scores for CoMMuTE translation slightly decrease after fine-tuning.
We attribute this to the dataset containing translations that stylistically differ.
Gemma-3-12B outperforms all other models on this task including the NLLB translation model.
Similarly, the Gemma-3 models are excellent at resolving lexical ambiguities on CoMMuTE followed by Pixtral-12B.
Our models fall behind, as they are trained on synthetic data and only a small dataset with real annotations.
The disambiguation accuracy stays constant for pre-training, while it increases with scale for fine-tuning.
MOF achieves 66.5\% mean accuracy from pre-training on web-crawled data but shows reduced Multi30K translation performance.
Our approach features 63.9\% mean accuracy while maintaining good results for Multi30K translation.
Overall, our findings highlight the need for small high-quality datasets with full task coverage in each language addressing these ambiguities, while the pre-training dataset can be of lower quality or even incomplete.

\section{Conclusion}
\label{sec:conclusion}

We presented scaling laws for generalization from multimodal machine translation to multilingual image captioning, demonstrating how transfer performance scales with the multilinguality of the base model, the model size, and the amount of training data.
While captioning in languages encountered only in the translation task still requires a language prefix in a zero-shot setting, our results highlight that these factors strongly influence how well encoder-decoder VLMs extend learned language capabilities to unseen task-language combinations.
Fine-tuning removes the need for explicit prefixes and yields competitive performance across downstream tasks.
Our insights can help practitioners create multilingual datasets more efficiently and make informed trade-offs between model size, multilingual pre-training, number of training samples, and task coverage.
Future work can investigate the interactions of more than two tasks and the extension of our findings to decoder-only VLMs, potentially leading to better, more versatile multilingual models.

\section*{Acknowledgments}
This research has been funded by the Federal Ministry of Education and Research of Germany under grant no. 01IS23004B RIDMI and 01IS22094C WEST-AI.
Computational resources were provided by the German AI Service Center WestAI.

\bibliography{aaai2026}

\clearpage

\appendix

\section{Preliminary Analysis}

We explore whether text-only machine translation models can resolve ambiguities with the help of additional textual context provided by a vision-language model (VLM).
We use CoMMuTE~\cite{futeral2023commute,futeral2025zerommt}, designed for binary disambiguation, featuring 155 ambiguous English sentences, each with two images and translations in German, French, Russian, and Chinese.
The textual context for each image is a detailed image description generated by \texttt{Llama3-LLaVA-NeXT-8B}\footnote{\label{footnote:llava_model}\url{https://hf.co/lmms-lab/llama3-llava-next-8b}}, a vision adaptation of Llama 3~\cite{dubey2024llama3,liu2023llava}.
The context is truncated to different lengths and is added after the caption, creating the input for NLLB~\cite{costajussa2022nllb}.
Based on perplexity, a measure of how well a probability distribution predicts a given sequence, the correct or incorrect translation with respect to the image is selected to calculate the accuracy.
With this training-free approach, lexical ambiguity can be resolved to some extent, as reported in Table~\ref{tab:text_only_commute}.
The best accuracy of 64.0\% is achieved by the larger NLLB-3.3B with an image description context length of about 64.

\begin{table}[b]
\small
\centering
\begin{tabularx}{\linewidth}{l*{5}{>{\centering\arraybackslash}X}}
\toprule
\multicolumn{1}{c}{} & \multicolumn{5}{c}{\textbf{Context Length}} \\
\cmidrule(lr){2-6}
\textbf{Model} & 0 & 32 & 64 & 128 & 256  \\
\midrule
NLLB-600M & 50.0 & 55.3 & 54.7 & 55.3 & 54.0 \\
NLLB-3.3B & 50.0 & 60.7 & \textbf{64.0} & 63.3 & 63.3 \\
\bottomrule
\end{tabularx}
\caption{Accuracy (\%) of binary disambiguation on the CoMMuTE En$\rightarrow$De subset~\cite{futeral2023commute} using the context-enhanced machine translation model NLLB~\cite{costajussa2022nllb} for various context lengths. Note that longer context and larger model result in better accuracy.}
\label{tab:text_only_commute}
\end{table}

\section{Model Overview}

The studied model is a standard encoder-decoder transformer composed from Florence-2~\cite{xiao2024florence2}, a series of small VLMs, and Gemma-2~\cite{google2024gemma2}, a large language model.
Florence-2 utilizes the DaViT~\cite{ding2022davit} image encoder.
The different model variants are detailed in Table~\ref{tab:florence2_models_sorted}.

\begin{table}[t]
    \centering
    \small
    \newcolumntype{Y}{>{\centering\arraybackslash}X}
    \begin{tabularx}{\linewidth}{l|>{\centering\arraybackslash}p{2.8cm}|Y|Y|Y}
        \toprule
        Param. & \textbf{0.4B} & \textbf{1B} & \textbf{3.5B} & \textbf{11.2B} \\
        \midrule
        & \multicolumn{4}{c}{\textbf{Vision Encoder}} \\
        & \multicolumn{4}{c}{DaViT} \\
        \midrule
        \# layers & [1, 1, 9, 1] & \multicolumn{3}{c}{[1, 1, 9, 1]} \\
        Hidden dim. & [128, 256, 512, 1024] & \multicolumn{3}{c}{[256, 512, 1024, 2048]} \\
        \# heads & [4, 8, 16, 32] & \multicolumn{3}{c}{[8, 16, 32, 64]} \\
        Patch size & [7, 3, 3, 3] & \multicolumn{3}{c}{[7, 3, 3, 3]} \\
        Patch stride & [4, 2, 2, 2] & \multicolumn{3}{c}{[4, 2, 2, 2]} \\
        \midrule
        & \multicolumn{4}{c}{\textbf{Encoder}} \\
        & \multicolumn{4}{c}{Florence-2} \\
        \midrule
        \# layers & 6 & \multicolumn{3}{c}{12} \\
        Hidden dim. & 768 & \multicolumn{3}{c}{1024} \\
        \# heads & 12 & \multicolumn{3}{c}{16} \\
        \midrule
        Vocab. size & \multicolumn{4}{c}{51328} \\
        \midrule
        & \multicolumn{4}{c}{\textbf{Decoder}} \\
        & \multicolumn{2}{c}{Florence-2} & \multicolumn{2}{c}{Gemma-2} \\
        \midrule
        \# layers & 6 & 12 & 26 & 42 \\
        Hidden dim. & 768 & 1024 & 2304 & 3584 \\
        \# heads & 12 & 16 & 8 & 16 \\
        \midrule
        Vocab. size & \multicolumn{4}{c}{256000} \\
        \bottomrule
    \end{tabularx}
    \caption{Model configuration.}
    \label{tab:florence2_models_sorted}
\end{table}

\section{Dataset Creation}

\begin{figure*}[t]
    \centering
    \includegraphics[width=\linewidth]{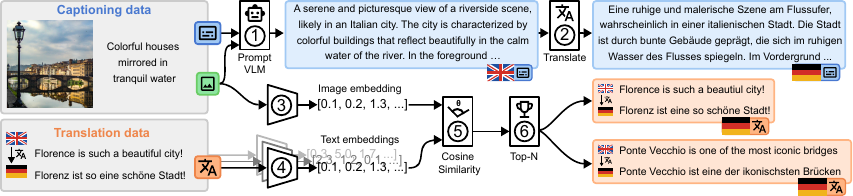}
    \caption{Dataset generation pipeline. Input is an image dataset with short captions or alt texts and a translation dataset with bitexts in English and the target language German. (1) Image and short caption are fed into a VLM to generate a detailed English description, which is (2) translated into the target language. (3) The image and (4) all English sentences of the translation dataset are embedded in a shared vector space. (5) Cosine similarity is calculated and (6) top-N matching pairs the most similar images and translations followed by deduplication.}
    \label{fig:data_pipeline}
\end{figure*}

For our investigation we propose a pipeline, illustrated in Figure~\ref{fig:data_pipeline}, for creating a multilingual multimodal corpus based on an image dataset and a translation dataset with parallel sentences, also known as bitexts.
We leverage machine translation models and contrastive models to improve semantic distinctiveness.
Our method addresses ambiguity in two ways: Improving caption translations by generating more synthetic textual context and linking parallel sentences with image context using contrastive models.
Short image captions~\cite{chen2015cococap,elliott2016multi30k} cannot provide enough context to resolve semantic ambiguities when translated.
Methods that use directly translated short image captions for training~\cite{thapliyal2022xm3600,qiu2022multilingual} may learn incorrect word mappings.
Therefore, we first generate detailed image descriptions using a VLM, as they provide extra information and improve training~\cite{liu2023llava,chen2024sharegpt4v}.
We translate the descriptions using a standard machine translation model, as they likely already provide enough context (see Table~\ref{tab:text_only_commute}).
To avoid information loss, we remove translations that have a different number of sentences than their source.
More sophisticated filters such as using token id statistics or BLEU between source and translation could potentially be applied as enhancements~\cite{qiu2022multilingual}.
The descriptions are further processed by extracting the first sentence as a short-form caption and combining the English description with the respective translation as a synthetic translation sample.
For translating original short-form captions, we use context-enhanced machine translation.
We append a detailed generated description with special characters as separation tokens to the short-from caption before translating.
Next, we check for those characters in the translation.
If they are missing, as is true in the majority of cases, we apply the machine translation models without context.
This process slightly enhances context awareness at the cost of translation accuracy.
Fine-tuning the translation model with a context prompt or leveraging LLMs as demonstrated by \citet{wang2024lambda} could potentially lower the high reject rate.

For our auxiliary translation task, we enrich the image annotations with a text-only machine translation dataset.
Contrastive models demonstrate impressive retrieval performance~\cite{radford2021clip}.
We encode the images and the English source sentence and calculate the cosine similarity between the resulting embeddings.
We pair bitexts with images with top-N matching, followed by deduplication. 
Faiss~\cite{douze2024faiss} can be employed to accelerate the matching process.
To create more sophisticated translation data, we randomly combine bitexts to create multi-sentence translation pairs.

\section{Datasets}

\begin{table*}[t]
\small
\centering
\newcolumntype{Y}{>{\centering\arraybackslash}X}
\begin{tabularx}{0.8\linewidth}{clclY}
\toprule
\textbf{Type} & \textbf{Dataset} & \textbf{Source} & \textbf{Languages} & \textbf{Train (Images~/~Texts)}\\ 
\midrule
\multirow{8}{*}{Caption} & COCO Karpathy & GT & En & 113,287 / 566,435 \\
& Multi30k Task 2 & GT & En, De & 29,000 / 290,000 \\
& Image Paragraphs & GT & En & 14,579 / 14,579 \\ 
& DOCCI & GT & En & 9,647 / 9,647  \\ 
\cmidrule(lr){2-5}
& Multi30k Task 2 & MT & Fr, Es, Ru, Zh & * / 580,000 \\
& Image Paragraphs & MT & De, Fr, Es, Ru, Zh & * / 72,895 \\ 
& DOCCI & MT & De, Fr, Es, Ru, Zh & * / 48,235 \\ 
\midrule
\multirow{2}{*}{%
Translation En$\rightarrow$X
} & Multi30k Task 1 & GT & De, Fr & 29,000 / 58,000 \\
\cmidrule(lr){2-5}
& Multi30k Task 1 & MT & Es, Ru, Zh & * / 87,000  \\
\bottomrule
\end{tabularx}
\caption{Fine-tuning dataset for the captioning and translation task based on COCO Karpathy~\cite{chen2015cococap,karpathy2017deep}, Multi30k~\cite{elliott2016multi30k}, Image Paragraphs~\cite{krause2017ip} and DOCCI~\cite{onoe2024docci}.The annotation is the regular ground truth (GT) or machine translated (MT).}
\label{tab:dataset_overview}
\end{table*}

\begin{figure}[t]
    \centering
    \includegraphics[trim={0 2.1cm 0 2.1cm},clip,width=\linewidth]{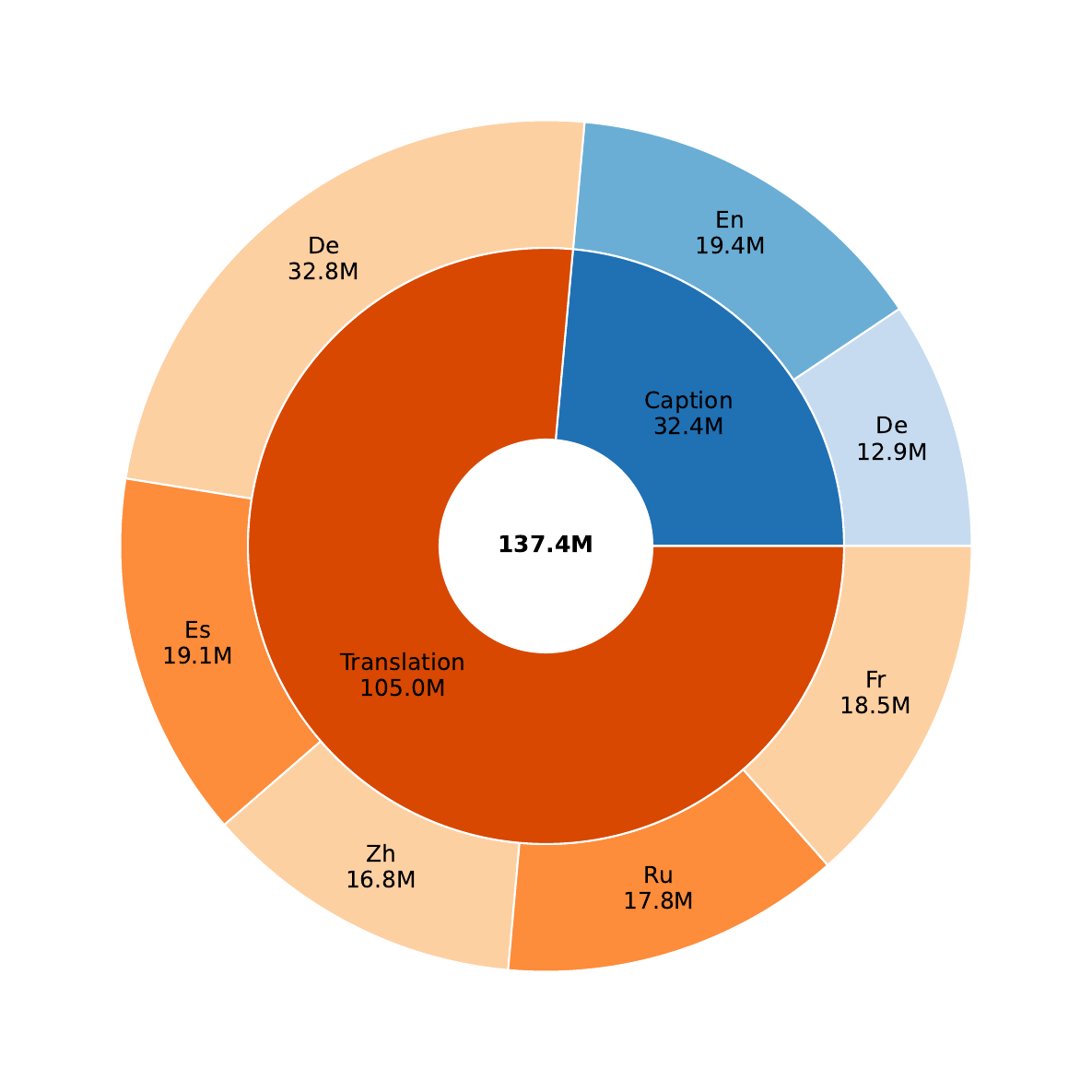}
    \caption{Distribution of caption and translation data with language coverage in our pre-training dataset.}
    \label{fig:dist_pretrain_dataset}
\end{figure}

\begin{figure}[t]
    \centering
    \includegraphics[trim={0 2cm 0 2cm},clip,width=\linewidth]{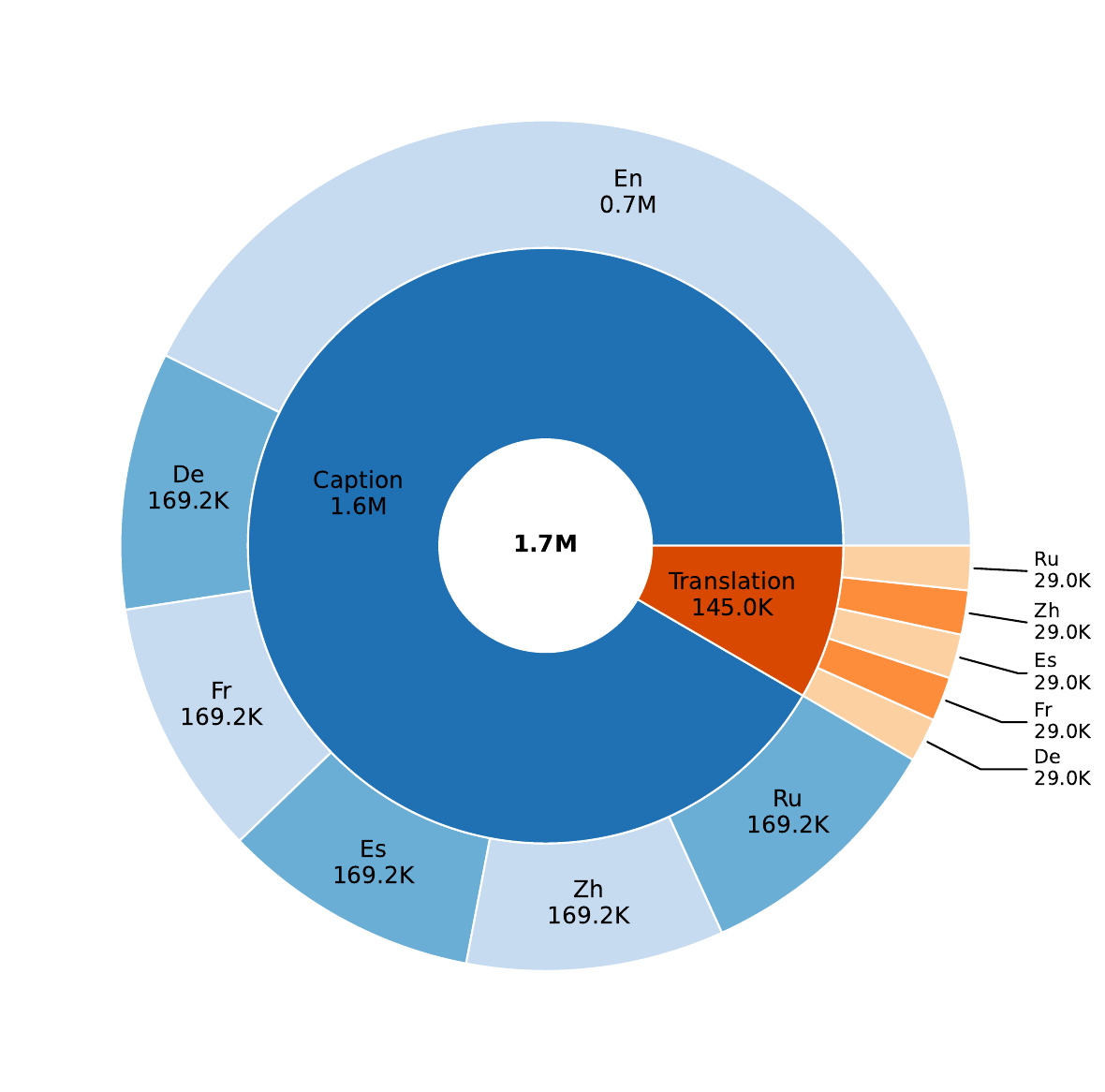}
    \caption{Distribution of caption and translation data with language coverage in our fine-tuning dataset.}
    \label{fig:dist_finetune_dataset}
\end{figure}

Figure~\ref{fig:dist_pretrain_dataset} shows the task and language coverage for our pre-training dataset.
The coverage for our fine-tuning dataset is illustrated in Figure~\ref{fig:dist_finetune_dataset}.
An overview of the dataset mix is shown in Table~\ref{tab:dataset_overview}.
We apply online task-language balancing, because of the high imbalance of captioning and translation data.

\section{Scaling Laws Details}

\subsection{Data}

The data collected for the scaling law study is detailed in Table~\ref{tab:data}.

\begin{table*}[ht]
\small
    \centering
\begin{tabularx}{0.7\linewidth}{l*{6}{>{\centering\arraybackslash}X}}
\toprule
\textbf{Param.} & \textbf{GMACs} & \textbf{Training samples} & \textbf{Initial loss} & \textbf{SC loss} & \textbf{ST loss} & \textbf{UC loss} \\
\midrule
0.4B & 62.1 & 0.5M & 10.44 & 1.92 & 4.20 & 5.11 \\
0.4B & 62.1 & 2.0M & 10.44 & 1.54 & 3.21 & 4.51 \\
0.4B & 62.1 & 5.1M & 10.44 & 1.38 & 2.81 & 4.37 \\
0.4B & 62.1 & 10.2M & 10.44 & 1.30 & 2.59 & 4.43 \\
\midrule
1.0B & 170.6 & 0.5M & 9.89 & 1.77 & 4.13 & 5.13 \\
1.0B & 170.6 & 5.1M & 9.89 & 1.26 & 2.75 & 4.17 \\
1.0B & 170.6 & 2.0M & 9.89 & 1.40 & 3.15 & 4.48 \\
1.0B & 170.6 & 10.2M & 9.89 & 1.17 & 2.49 & 4.15 \\
\midrule
3.5B & 488.8 & 0.5M & 5.85 & 1.57 & 3.56 & 3.44 \\
3.5B & 488.8 & 2.0M & 5.85 & 1.15 & 2.15 & 3.15 \\
3.5B & 488.8 & 5.1M & 5.85 & 1.03 & 1.95 & 3.10 \\
3.5B & 488.8 & 10.2M & 5.85 & 0.97 & 1.86 & 3.14 \\
\midrule
11.2B & 1494.5 & 0.5M & 6.02 & 1.34 & 3.14 & 3.04 \\
11.2B & 1494.5 & 2.0M & 6.02 & 1.01 & 1.88 & 2.85 \\
11.2B & 1494.5 & 5.1M & 6.02 & 0.92 & 1.72 & 2.87 \\
11.2B & 1494.5 & 10.2M & 6.02 & 0.87 & 1.65 & 2.88 \\
\bottomrule
\end{tabularx}
    \caption{Data collected for the scaling law study.}
    \label{tab:data}
\end{table*}

\subsection{OLS Diagnostics and Fit Robustness}

Diagnostics to test OLS assumptions for the second power law (Equation~\ref{eq:powerlaw2}) are reported in Table~\ref{tab:ols_diagnostics}.
An extension of Table~\ref{tab:beta_parameters} is given in Table~\ref{tab:beta_parameters_extended} with 95\% CIs derived with HC3 robust standard errors and bootstrapping.
For bootstrapping we randomly sample rows of the dataset with replacement and recalculate the OLS model 10{,}000 times.
Model fit is summarized in Table~\ref{tab:model_fit_metrics}.

\begin{table*}[t]
\centering
\begin{tabularx}{0.75\linewidth}{>{\raggedright\arraybackslash}Xcccccc}
\toprule
\textbf{Test} & \multicolumn{2}{c}{SC} & \multicolumn{2}{c}{ST} & \multicolumn{2}{c}{UC} \\
\cmidrule(lr){2-7}
 & Statistic & p-value & Statistic & p-value & Statistic & p-value \\
\midrule
Condition Number & 4.611 &  & 4.611 &  & 4.611 &  \\
Breusch-Pagan (LM) & 5.628 & 0.131 & 6.463 & 0.091 & 2.978 & 0.395 \\
Breusch-Pagan (F) & 2.170 & 0.145 & 2.711 & 0.092 & 0.915 & 0.463 \\
Durbin-Watson & 2.098 &  & 1.872 &  & 1.848 &  \\
Ramsey RESET & 0.015 & 0.904 & 0.171 & 0.687 & 0.121 & 0.734 \\
Jarque-Bera & 0.357 & 0.836 & 0.354 & 0.838 & 0.918 & 0.632 \\
Omnibus & 0.164 & 0.921 & 0.156 & 0.925 & 2.252 & 0.324 \\
\bottomrule
\end{tabularx}
\caption{OLS diagnostic tests for the second power law (Equation~\ref{eq:powerlaw2}).}
\label{tab:ols_diagnostics}
\end{table*}

\begin{table*}[t]
\centering
\begin{tabularx}{0.85\linewidth}{Xcccccc}
\toprule
\textbf{Task} & \textbf{Coefficient} & \textbf{Estimate} & \textbf{OLS 95\% CI} & \textbf{HC3 95\% CI} & \textbf{Bootstrap 95\% CI} & $\bm{p}$\textbf{-value} \\
\midrule
\multirow{3}{*}{SC} & $\beta_1$ & -0.59 & [-0.79, -0.38] & [-0.85, -0.33] & [-0.79, -0.40] & $\bm{p} < 0.001$ \\
& $\beta_2$ & -0.72 & [-0.80, -0.63] & [-0.82, -0.61] & [-0.79, -0.61] & $\bm{p} < 0.001$ \\
& $\beta_3$ & 0.10 & [-0.10, 0.31] & [-0.17, 0.38] & [-0.11, 0.29] & $\bm{p} = 0.293$ \\
\midrule
\multirow{3}{*}{ST} & $\beta_1$ & -0.36 & [-0.74, 0.03] & [-0.85, 0.14] & [-0.69, 0.05] & $\bm{p} = 0.068$ \\
& $\beta_2$ & -0.73 & [-0.89, -0.57] & [-0.92, -0.54] & [-0.85, -0.52] & $\bm{p} < 0.001$ \\
& $\beta_3$ & 0.29 & [-0.09, 0.68] & [-0.20, 0.78] & [-0.03, 0.71] & $\bm{p} = 0.125$ \\
\midrule
\multirow{3}{*}{UC} & $\beta_1$ & -0.41 & [-0.69, -0.12] & [-0.69, -0.13] & [-0.58, -0.06] & $\bm{p} = 0.009$ \\
& $\beta_2$ & -0.23 & [-0.35, -0.11] & [-0.38, -0.08] & [-0.35, -0.10] & $\bm{p} = 0.001$ \\
& $\beta_3$ & 0.57 & [0.29, 0.85] & [0.29, 0.85] & [0.38, 0.86] & $\bm{p} < 0.001$ \\
\bottomrule
\end{tabularx}
\caption{Extended version of Table~\ref{tab:beta_parameters}. We additionally report HC3 robust 95\% CI and bootstrap percentile 95\% CI.}
\label{tab:beta_parameters_extended}
\end{table*}

\begin{table}[t]
\small
\centering
\begin{tabularx}{0.8\linewidth}{Xccc}
\toprule
\textbf{Task} & $\bm{R^2}$ & $\bm{R^2_{\text{adj}}}$ & $\bm{\text{LOOCV }R^2}$ \\
\midrule
SC & 0.982 & 0.977 & 0.965 \\
ST & 0.935 & 0.919 & 0.873 \\
UC & 0.965 & 0.956 & 0.936 \\
\bottomrule
\end{tabularx}
\caption{Model fit metrics for the second power law in log$_{10}$ space: $R^2$, adjusted $R^2$, and leave-one-out cross-validated $R^2$.}
\label{tab:model_fit_metrics}
\end{table}

\section{Evaluation for Downstream Tasks}
\label{a:section:eval}

In this section we describe the evaluation strategy for the individual tasks.

\noindent
\textbf{COCO Karpathy evaluation.}
We use the \texttt{pycocoevalcap} toolkit\footnote{\url{https://github.com/salaniz/pycocoevalcap}} to calculate BLEU~\cite{papineni2002bleu}, METEOR-1.5~\cite{banerjee2005meteor}, and CIDEr~\cite{vedantam2015cider}.
This is the regular procedure following related work~\cite{li2023blip2}, where CIDEr is considered the main metric. 
Moreover, we report the neural metric CLIPScore (CS)~\cite{hessel2021clipscore} with CLIP-ViT-B/16~\cite{radford2021clip} to measure alignment between generated captions and images in the CLIP embedding space.

\noindent
\textbf{Multi30K translation evaluation}
For the translation task (task 1) of Multi30K we follow related work~\cite{futeral2025zerommt} and use BLEU~\cite{papineni2002bleu} and COMET with the \texttt{wmt22-comet-da}\footnote{\url{https://hf.co/Unbabel/wmt22-comet-da}} model (C$_{22}$)~\cite{rei2020comet,rei2022comet22}.
BLEU is evaluated using SacreBLEU~\cite{post2019sacrebleu} with the \texttt{13a} tokenizer and for Chinese the \texttt{zh} tokenizer.

\noindent
\textbf{Multi30K captioning evaluation}
For the captioning task (task 2) of Multi30K, the same strategy as for COCO Karpathy is applied.
For multilingual tasks we use the multilingual CLIPScore (MS) with M-CLIP-XLM-R-L-ViT-B/16+~\cite{carlsson2022mclip}.

\begin{table}[t]
\small
\centering
\newcolumntype{Y}{>{\centering\arraybackslash}X}
\begin{tabularx}{\linewidth}{lYYYYYY}
\toprule
\textbf{Model} & \textbf{En} & \textbf{De} & \textbf{Fr} & \textbf{Es} & \textbf{Ru} & \textbf{Zh} \\
\midrule
Paper & 78.0 & - & - & - & - & - \\
Reprod. & 78.4 & 35.6 & 63.6 & 68.0 & 35.3 & 28.4 \\
\texttt{stanza} & 78.7 & 35.9 & 63.2 & 67.9 & 35.6 & 28.2 \\
Baseline-6B & 81.9 & 38.1 & 64.9 & 63.9 & 29.4 & 21.5 \\
\bottomrule
\end{tabularx}
\caption{Comparison between the numbers reported in the PaliGemma paper~\cite{beyer2024paligemma}, reproduction with pre-segmented model outputs, reproduction with \texttt{stanza}~\cite{qi2020stanza} for word segmentation, and the Baseline-6B~\cite{li2023blip2,costajussa2022nllb} for reference. Note that the outputs of PaliGemma-3B are pre-segmented, with words separated by whitespace. For the \texttt{stanza} evaluation, we first remove the whitespace before processing.}
\label{tab:xm3600_stanza}
\end{table}

\noindent
\textbf{XM3600 evaluation}
The main metric for XM3600~\cite{thapliyal2022xm3600} is CIDEr~\cite{vedantam2015cider}.
However, the pre-processing steps are not well documented for languages without word boundaries such as Chinese.
\citet{chen2023pali} uses a neural model to segment languages without word boundaries.
However, the exact model is not specified, thus making reproduction difficult.
Furthermore, they fine-tune their models on the pre-tokenized dataset COCO35L~\cite{thapliyal2022xm3600}, COCO captions~\cite{chen2015cococap} translated into 35 target languages.
PaliGemma~\cite{beyer2024paligemma,steiner2024paligemma2} follows a similar setup and is publicly available. 
These models produce segmented lower-case text, not requiring additional segmentation for evaluation.

For correct capitalization and white space, \citet{futeral2024moscar} proposes \texttt{stanza}~\cite{qi2020stanza} as an alternative for the unknown word segmentation network.
We apply the model both to predictions and to references and find that \texttt{stanza} is a good alternative, as can be seen in Table~\ref{tab:xm3600_stanza}.
Whether this approximation holds for languages other than Chinese, where word boundaries are not defined, is left as future work.

\section{Detailed Downstream Tasks Results}
We provide detailed results for the benchmarks: Multi30K (task 1) translation in Table~\ref{tab:multi30k}, CoMMuTE translation in Table~\ref{tab:commute_translation}, CoMMuTE disambiguation in Table~\ref{tab:lexical_ambiguity_commpute}, COCO Karpathy image captioning in Table~\ref{tab:coco_karpathy}, Multi30K (task 2) image captioning in Table~\ref{tab:multi30k_ic}, and XM3600 image captioning in Table~\ref{tab:xm3600_ic}.

\section{Qualitative Results}
\begin{figure*}[ht]
    \centering
    \includegraphics[width=0.95\linewidth]{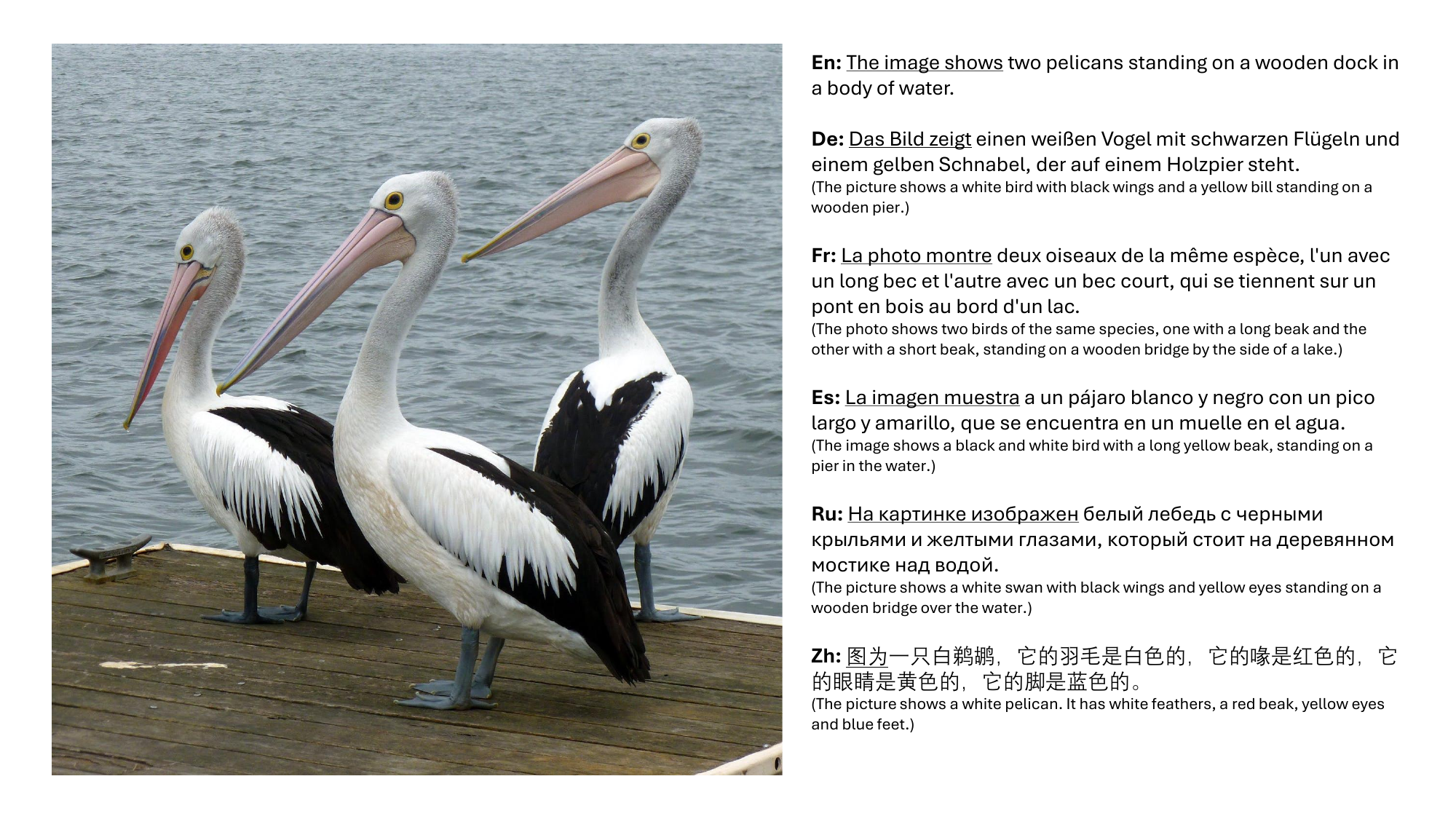}
    \caption{Captions generated with 11.2B by prompting with the \underline{underlined prefix}. 11.2B has not seen captioning data for Fr, Es, Ru, and Zh. English references are created with DeepL. Photo by Jeanette Dickson.}
    \label{fig:sample1}
\end{figure*}

\begin{figure*}[ht]
    \centering
    \includegraphics[width=0.95\linewidth]{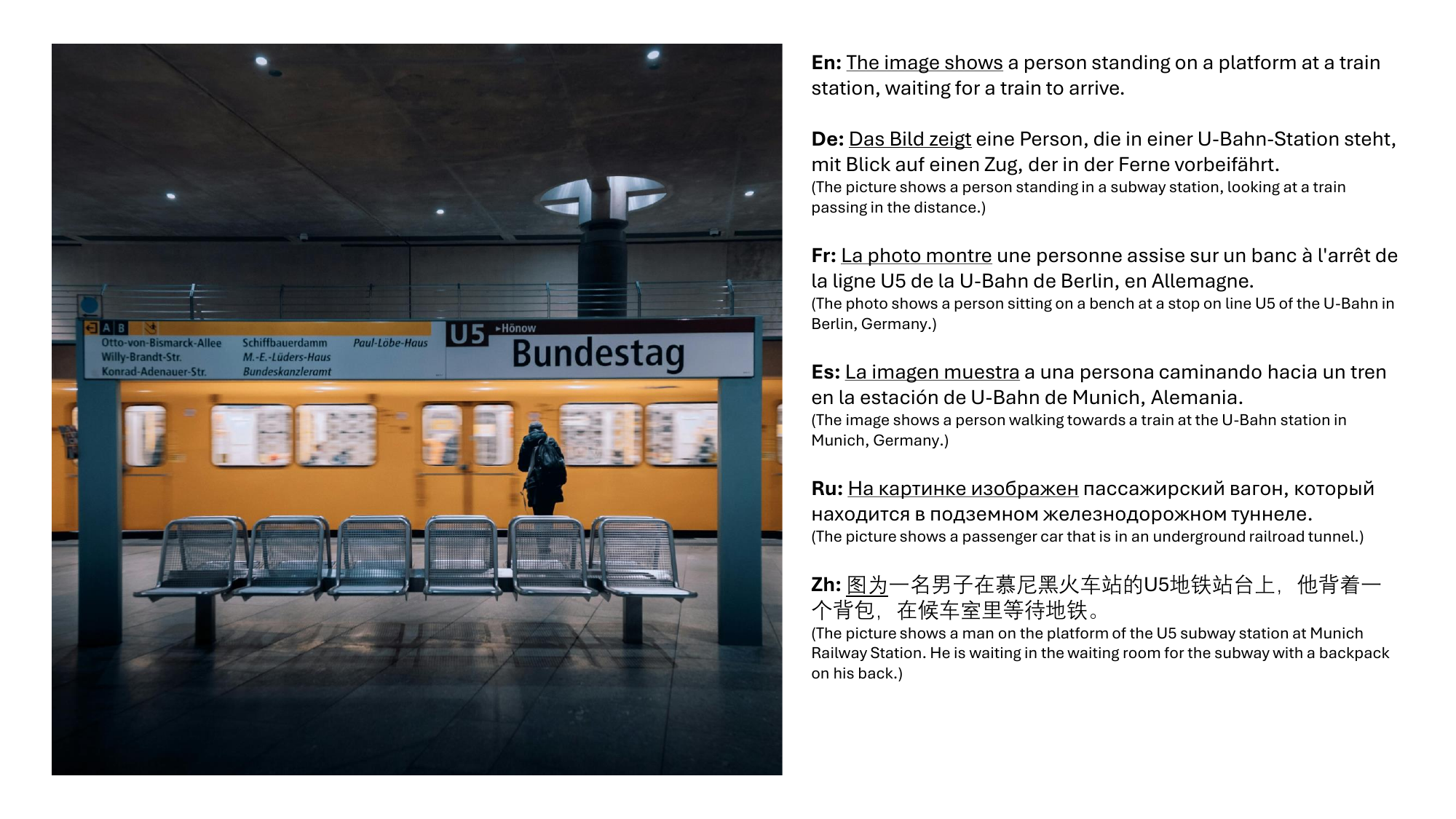}
    \caption{Captions generated with 11.2B by prompting with the \underline{underlined prefix}. 11.2B has not seen captioning data for French (Fr), Spanish (Es), Russian (Ru), and Chinese (Zh). English references are created with DeepL. Photo by Reinaldo Simoes.}
    \label{fig:sample2}
\end{figure*}

\begin{figure*}[ht]
    \centering
    \includegraphics[width=0.95\linewidth]{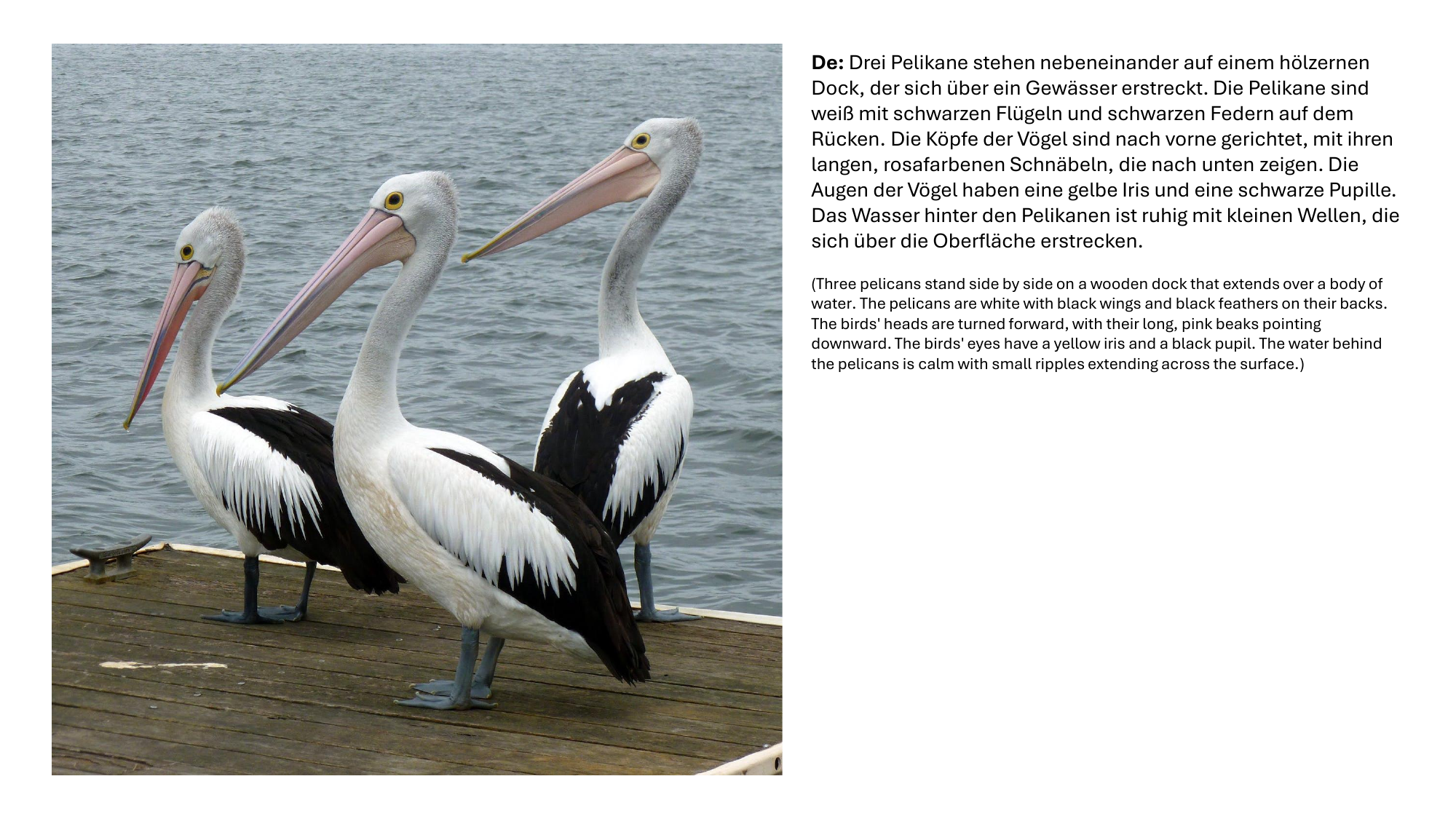}
    \caption{Detailed caption generated with 11.2B \textit{30K ft} for German (De). English reference is created with DeepL. Photo by Jeanette Dickson.}
    \label{fig:sample3}
\end{figure*}

\begin{figure*}[ht]
    \centering
    \includegraphics[width=0.95\linewidth]{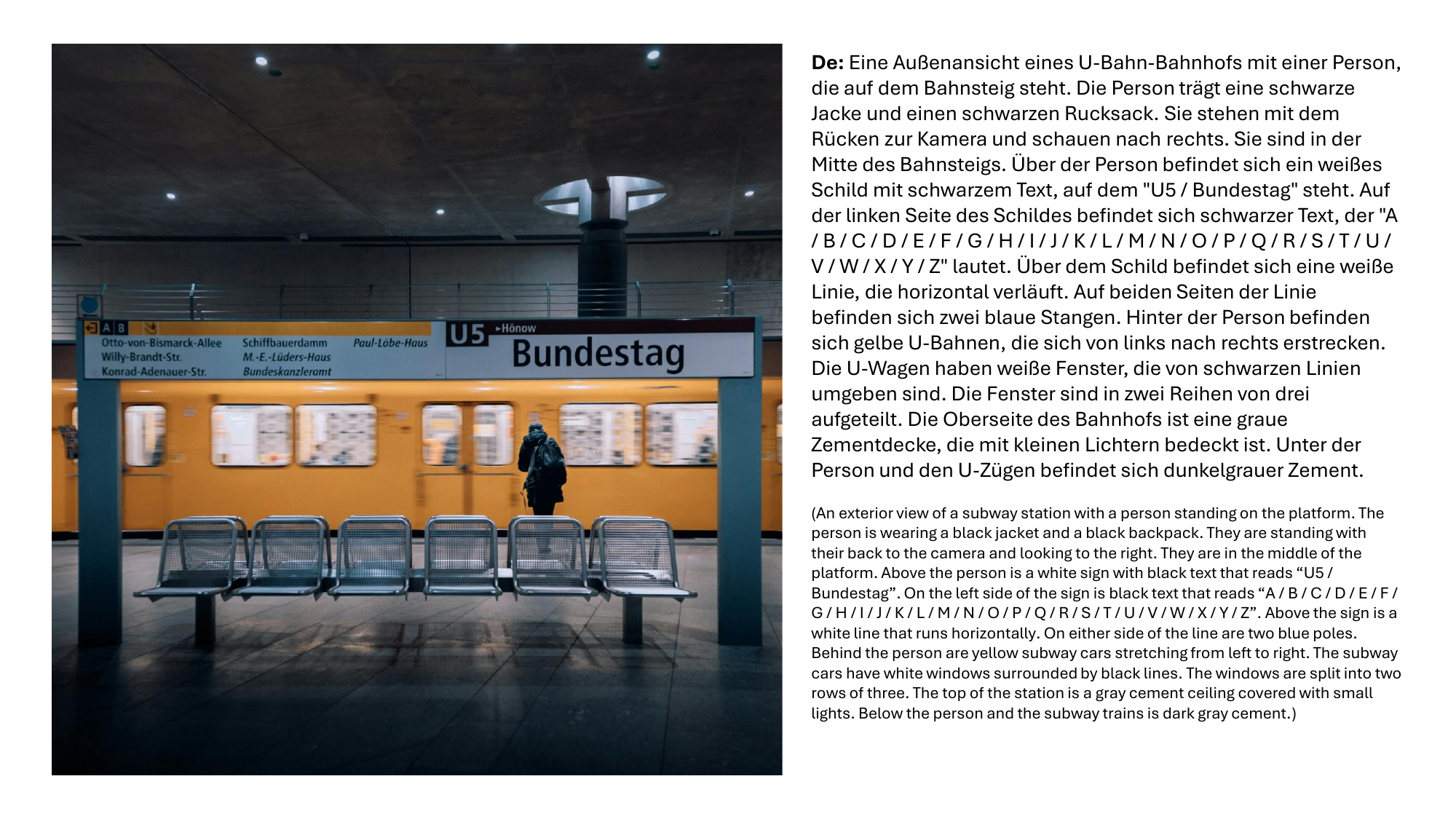}
    \caption{Detailed caption generated with 11.2B \textit{30K ft} for German (De). English reference is created with DeepL. Photo by Reinaldo Simoes.}
    \label{fig:sample4}
\end{figure*}

We provide example outputs for our best performing pre-trained model 11.2B and our best performing fine-tuned model 11.2B pre-trained for 30K steps. 
Figure~\ref{fig:sample1} and Figure~\ref{fig:sample2} show the outputs of the pre-trained model with examples for captioning in languages, where no captioning data was encountered.
Figure~\ref{fig:sample3} and Figure~\ref{fig:sample4} show detailed captions.

\begin{table*}[ht]
\small
    \centering
    \newcolumntype{Y}{>{\centering\arraybackslash}X}
    \begin{tabularx}{0.9\linewidth}{lYYYYYYYY}
        \toprule
        \multicolumn{9}{c}{\textbf{Multi30K Translation En$\rightarrow$Fr}} \\
        \midrule
         & \multicolumn{2}{c}{\textbf{Test2016}} & \multicolumn{2}{c}{\textbf{Test2017}} & \multicolumn{2}{c}{\textbf{Test2018}} & \multicolumn{2}{c}{\textbf{COCO}} \\
        \cmidrule(lr){2-3} \cmidrule(lr){4-5} \cmidrule(lr){6-7} \cmidrule(lr){8-9}
         \textbf{Model} & B & C$_{22}$ & B & C$_{22}$ & B & C$_{22}$ & B & C$_{22}$  \\
        \midrule
        Gemma-3-12B & 52.8 & 87.9 & 53.6 & 88.5 & 44.1 & 85.0 & 50.2 & 85.6 \\
        Gemma-3-4B & 49.6 & 86.9 & 48.3 & 86.5 & 41.7 & 84.1 & 46.0 & 84.2 \\
        Pixtral-12B & 54.6 & 87.4 & 52.8 & 87.6 & 42.9 & 84.3 & 54.0 & 85.3 \\
        \midrule
        VGAMT & 65.0 & 88.7 & 58.9 & 88.2 &  &  & 51.2 & 84.8 \\
        MOF & 36.0 & 83.6 & 35.1 & 83.7 &  &  & 34.1 & 80.7 \\
        \midrule
        ZeroMMT-600M & 48.6 & 84.9 & 48.1 & 85.7 &  &  & 50.3 & 83.8 \\
        ZeroMMT-1.3B & 51.5 & 86.4 & 51.1 & 87.0 &  &  & 53.6 & 85.0 \\
        ZeroMMT-3.3B & 52.9 & 87.2 & 53.3 & 87.5 &  &  & 53.9 & 85.4 \\
        \midrule
        NLLB-600M & 48.3 & 85.0 & 48.5 & 85.9 & 40.5 & 83.0 & 50.0 & 84.1 \\
        NLLB-1.3B & 51.8 & 86.6 & 50.9 & 86.9 & 43.0 & 84.1 & 52.8 & 85.0 \\
        NLLB-3.3B & 54.2 & 87.6 & 52.8 & 87.5 & 45.4 & 84.8 & 54.1 & 85.6 \\
        Baseline-6B & 54.8 & 87.6 & 53.8 & 87.6 & 45.7 & 84.8 & 54.4 & 85.6 \\
        \midrule
        0.4B \textit{ft} (ours) & 57.5 & 85.9 & 53.5 & 85.9 & 39.6 & 82.3 & 50.0 & 82.4 \\
        0.4B \textit{100k ft} (ours) & 60.4 & 87.7 & 57.5 & 87.9 & 43.0 & 84.0 & 53.8 & 84.7\\
        1.0B \textit{ft} (ours) & 59.9 & 87.5 & 56.9 & 87.3 & 42.4 & 83.5 & 52.9 & 84.2 \\
        3.5B \textit{ft} (ours) & 59.9 & 88.5 & 57.5 & 88.4 & 42.2 & 84.2 & 53.5 & 85.2 \\
        11.2B \textit{ft} (ours) & 63.2 & 89.2 & 60.3 & 89.1 & 43.9 & 84.9 & 53.9 & 85.8 \\
        11.2B \textit{30K ft} (ours) & 64.4 & 89.5 & 60.5 & 89.0 & 44.6 & 85.1 & 53.8 & 86.1 \\
        \midrule
        0.4B (ours) & 42.1 & 81.6 & 43.4 & 82.1 & 36.2 & 79.6 & 47.6 & 80.4 \\
        0.4B \textit{100K} (ours) & 49.0 & 85.0 & 50.1 & 85.9 & 42.2 & 83.1 & 53.9 & 83.7 \\
        1.0B (ours) & 45.3 & 83.6 & 46.3 & 84.4 & 39.1 & 82.0 & 50.7 & 82.0 \\
        3.5B (ours) & 46.5 & 85.1 & 48.0 & 85.8 & 40.3 & 83.1 & 50.5 & 83.1 \\
        11.2B (ours) & 50.6 & 86.0 & 50.2 & 86.7 & 42.2 & 83.6 & 52.0 & 84.4 \\
        11.2B \textit{30K} (ours) & 51.3 & 86.1 & 51.4 & 87.0 & 43.4 & 83.8 & 54.3 & 84.9 \\
        \midrule
        \multicolumn{9}{c}{\textbf{Multi30K Translation En$\rightarrow$De}} \\
        \midrule
        Gemma-3-12B & 43.0 & 87.1 & 38.8 & 86.6 & 36.4 & 85.3 & 35.8 & 83.3 \\
        Gemma-3-4B & 39.8 & 85.4 & 34.7 & 84.6 & 33.2 & 83.4 & 31.9 & 80.7 \\
        Pixtral-12B & 41.4 & 86.6 & 37.5 & 86.0 & 34.8 & 84.3 & 34.9 & 82.6 \\
        \midrule
        VGAMT & 41.9 & 85.8 & 36.7 & 84.7 &  &  & 33.5 & 81.1 \\
        MOF & 28.9 & 82.3 & 23.9 & 80.9 &  &  & 21.9 & 76.6 \\
         \midrule
        ZeroMMT-600M & 36.2 & 83.0 & 33.2 & 82.5 &  &  & 29.0 & 77.7 \\
        ZeroMMT-1.3B & 37.6 & 84.0 & 36.2 & 84.6 &  &  & 31.7 & 80.7 \\
        ZeroMMT-3.3B & 39.6 & 85.9 & 37.9 & 85.5 &  &  & 33.7 & 81.9 \\
        \midrule
        NLLB-600M & 37.2 & 83.5 & 33.0 & 83.0 & 32.2 & 81.9 & 28.3 & 78.6 \\
        NLLB-1.3B & 38.1 & 85.0 & 36.7 & 84.3 & 32.8 & 83.2 & 31.5 & 80.4 \\
        NLLB-3.3B & 39.7 & 86.1 & 37.9 & 85.2 & 35.8 & 84.2 & 34.5 & 82.0 \\
        Baseline-6B & 39.8 & 86.2 & 38.3 & 85.5 & 35.6 & 84.3 & 33.9 & 82.1 \\
        \midrule
        0.4B \textit{ft} (ours) & 42.8 & 84.5 & 37.8 & 84.0 & 34.6 & 82.4 & 32.4 & 79.6 \\
        0.4B \textit{100K ft} (ours) & 44.0 & 86.3 & 40.9 & 86.1 & 36.5 & 84.1 & 34.8 & 82.1  \\
        1.0B \textit{ft} (ours) & 43.2 & 85.9 & 38.7 & 85.3 & 35.7 & 83.7 & 33.3 & 80.8 \\
        3.5B \textit{ft} (ours) & 43.1 & 86.6 & 39.8 & 85.9 & 36.8 & 84.4 & 34.7 & 82.0 \\
        11.2B \textit{ft} (ours) & 44.7 & 87.3 & 41.6 & 87.0 & 37.7 & 85.4 & 35.7 & 83.6 \\
        11.2B \textit{30K ft} (ours) & 44.9 & 87.6 & 41.6 & 87.0 & 38.8 & 85.6 & 38.5 & 83.9 \\
        \midrule
        0.4B (ours) & 37.2 & 83.1 & 35.9 & 83.6 & 32.9 & 81.9 & 29.3 & 78.4 \\
        0.4B \textit{100K} (ours) & 39.5 & 85.0 & 37.1 & 84.7 & 34.7 & 83.1 & 31.8 & 80.5 \\
        1.0B (ours) & 38.8 & 84.3 & 36.3 & 84.3 & 34.6 & 83.0 & 30.6 & 79.1 \\
        3.5B (ours) & 38.2 & 84.7 & 37.2 & 84.6 & 34.5 & 83.4 & 31.8 & 80.8 \\
        11.2B (ours) & 39.3 & 85.7 & 37.7 & 85.4 & 34.6 & 83.9 & 32.7 & 81.1 \\
        11.2B \textit{30K} (ours) & 39.5 & 85.8 & 38.3 & 85.4 & 36.0 & 84.1 & 35.0 & 81.6 \\
        \bottomrule
    \end{tabularx}
    \caption{Translation (Task 1) on the Multi30k~\cite{elliott2016multi30k,elliott2017multi30ktask1test2017,barrault2018multi30ktest2018} dataset reporting BLEU (B) and COMET (C$_{22}$) metrics. \textit{100K}, \textit{30K}, and \textit{ft} indicate the number of pre-training steps and fine-tuning.}
    \label{tab:multi30k}
\end{table*}

\begin{table*}[ht]
\small
    \centering
    \begin{tabularx}{0.7\linewidth}{l*{8}{>{\centering\arraybackslash}X}}
        \toprule
        & \multicolumn{2}{c}{\textbf{De}} & \multicolumn{2}{c}{\textbf{Fr}} & \multicolumn{2}{c}{\textbf{Ru}} & \multicolumn{2}{c}{\textbf{Zh}} \\
        \cmidrule(lr){2-3} \cmidrule(lr){4-5} \cmidrule(lr){6-7} \cmidrule(lr){8-9}
        \textbf{Model} & B & C$_{22}$ & B & C$_{22}$ & B & C$_{22}$ & B & C$_{22}$ \\
        \midrule
        Gemma-3-12B & 44.1 & 85.6 & 45.2 & 84.7 & 29.1 & 86.6 & 35.5 & 85.0 \\
        Gemma-3-4B & 35.6 & 83.6 & 40.0 & 82.2 & 27.5 & 84.9 & 17.4 & 49.0 \\
        Pixtral-12B & 40.7 & 84.8 & 45.9 & 83.7 & 23.5 & 84.1 & 33.5 & 84.2 \\
        \midrule
        NLLB-600M &  36.4 & 80.6 & 38.9 & 80.2 & 19.8 & 80.1 & 20.8 & 75.5 \\
        NLLB-1B & 40.5 & 81.3 & 39.2 & 80.4 & 23.7 & 81.9 & 20.3 & 75.6 \\
        NLLB-3.3B & 40.8 & 81.3 & 41.4 & 81.4 & 22.7 & 82.4 & 22.8 & 77.0 \\
        Baseline-6B & 41.5 & 81.6 & 42.1 & 81.5 & 23.5 & 82.9 & 23.0 & 77.2 \\
        \midrule
        0.4B \textit{ft} (ours) & 26.5 & 71.6 & 27.8 & 72.9 & 9.9 & 69.2 & 19.1 & 73.1 \\
        0.4B \textit{100K ft} (ours) & 32.8 & 78.3 & 37.6 & 78.6 & 13.6 & 77.5 & 22.5 & 75.0 \\
        1.0B \textit{ft} (ours) & 29.0 & 74.3 & 30.5 & 75.4 & 12.0 & 72.7 & 19.5 & 74.3 \\
        3.5B \textit{ft} (ours) & 32.0 & 79.3 & 35.5 & 78.6 & 15.0 & 77.9 & 21.5 & 75.2 \\
        11.2B \textit{ft} (ours) & 36.8 & 81.5 & 40.7 & 81.3 & 18.1 & 80.9 & 22.7 & 75.4 \\
        11.2B \textit{30K ft} (ours) & 38.5 & 82.4 & 42.6 & 81.9 & 19.6 & 81.3 & 23.9 & 76.6  \\
        \midrule
        0.4B (ours) & 34.1 & 77.5 & 31.0 & 76.4 & 12.5 & 71.1 & 26.2 & 76.6 \\
        0.4B \textit{100K} (ours) & 37.2 & 80.4 & 37.4 & 78.5 & 16.4 & 77.7 & 27.1 & 77.8 \\
        1.0B (ours) & 35.4 & 78.8 & 32.4 & 77.0 & 14.2 & 75.5 & 26.2 & 77.1 \\
        3.5B (ours) & 36.7 & 80.0 & 36.8 & 78.1 & 14.6 & 79.1 & 26.9 & 77.7 \\
        11.2B (ours) & 39.5 & 80.7 & 37.5 & 79.1 & 16.1 & 79.2 & 26.3 & 76.6 \\
        11.2B \textit{30K} (ours) & 39.4 & 80.5 & 35.8 & 78.8 & 17.6 & 78.8 & 28.6 & 77.4 \\
        \bottomrule
    \end{tabularx}
    \caption{Translation results on CoMMuTE~\cite{futeral2023commute,futeral2025zerommt} reporting BLEU (B) and COMET (C$_{22}$) metrics for German (De), French (Fr), Russian (Ru), and Chinese (Zh). \textit{100K}, \textit{30K}, and \textit{ft} indicate the number of pre-training steps and fine-tuning.}
    \label{tab:commute_translation}
\end{table*}

\begin{table*}[ht]
\small
    \centering
    \begin{tabularx}{0.6\linewidth}{l*{4}{>{\centering\arraybackslash}X}}
        \toprule
        \textbf{Model} & \textbf{De} & \textbf{Fr} & \textbf{Ru} & \textbf{Zh} \\
        \midrule
        Gemma-3-12B & 73.3 & 79.5 & 76.5 & 77.2 \\
        Gemma-3-4B & 74.7 & 76.6 & 74.7 & 74.1 \\
        Pixtral-12B & 73.3 & 76.3 & 76.9 & 75.6 \\
        \midrule
        MOF & 63.7 & 68.5 & 67.4 & 66.5 \\
        Baseline-6B & 61.7 & 62.0 & 58.6 & 62.0 \\
        \midrule
        0.4B \textit{ft} (ours) & 59.0 & 55.8 & 53.4 & 58.3 \\
        0.4B \textit{100K ft} (ours) & 59.7 & 62.3 & 55.6 & 61.4 \\
        1.0B \textit{ft} (ours) & 59.7 & 60.7 & 55.2 & 59.6 \\
        3.5B \textit{ft} (ours) & 61.3 & 65.3 & 62.0 & 61.7 \\
        11.2B \textit{ft} (ours) & 62.3 & 67.9 & 63.6 & 61.7 \\
        11.2B \textit{30K ft} (ours) & 62.0 & 67.5 & 62.3 & 61.7 \\
        \midrule
        0.4B (ours) & 54.0 & 53.9 & 51.5 & 54.9 \\
        0.4B \textit{100K} (ours) & 54.0 & 56.5 & 51.2 & 54.9 \\
        1.0B (ours) & 54.7 & 54.5 & 51.9 & 55.2 \\
        3.5B (ours) & 53.0 & 54.9 & 54.0 & 55.2 \\
        11.2B (ours) & 52.7 & 52.9 & 54.0 & 54.3 \\
        11.2B \textit{30K} (ours) & 52.0 & 54.5 & 52.8 & 53.4 \\
        \bottomrule
    \end{tabularx}
    \caption{Binary disambiguation results on CoMMuTE~\cite{futeral2023commute,futeral2025zerommt} (accuracy in \%) for the En$\rightarrow$X translation setting. \textit{100K}, \textit{30K}, and \textit{ft} indicate the number of pre-training steps and fine-tuning.}
    \label{tab:lexical_ambiguity_commpute}
\end{table*}

\begin{table*}[ht]
\small
    \centering
    \begin{tabularx}{0.6\linewidth}{l*{3}{>{\centering\arraybackslash}X}}
        \toprule
        & B@4 & C & CS \\
        \midrule
        Gemma-3-12B & 15.9 & 48.1 & 82.2 \\
        Gemma-3-4B & 12.3 & 40.9 & 83.8 \\
        Pixtral-12B & 9.2 & 61.1 & 73.2 \\
        \midrule
        PaliGemma-3B & 41.6 & 141.7 & 77.1 \\
        Florence-2-base &  & 140.0 &  \\
        Florence-2-large &  & 143.3 &  \\
        \midrule
        Baseline-6B & 42.8 & 145.2 & 76.9 \\
        \midrule
        0.4B \textit{ft} (ours) & 40.6 & 138.2 & 76.2 \\
        0.4B \textit{100K ft} (ours) & 41.3 & 140.3 & 76.6 \\
        1.0B \textit{ft} (ours) & 41.2 & 142.8 & 77.4 \\
        3.5B \textit{ft} (ours) & 41.8 & 141.6 & 76.6 \\
        11.2B \textit{ft} (ours) & 41.7 & 141.3 & 76.6 \\
        11.2B \textit{30K ft} (ours) & 41.6 & 140.8 & 76.6 \\
        \midrule
        0.4B (ours) & 13.2 & 28.2 & 77.7 \\
        0.4B \textit{100K} (ours) & 12.7 & 27.3 & 77.9 \\
        1.0B (ours) & 12.6 & 21.3 & 79.9 \\
        3.5B (ours) & 13.3 & 28.7 & 78.5 \\
        11.2B (ours) & 13.7 & 30.5 & 78.9 \\
        11.2B  \textit{30K} (ours) & 15.1 & 24.9 & 79.2  \\
        \bottomrule
    \end{tabularx}
    \caption{Image captioning results on COCO Karpathy~\cite{chen2015cococap,karpathy2017deep} test split for English reporting the metrics BLEU-4 (B@4), CIDEr (C) and CLIPScore (CS). \textit{100K}, \textit{30K}, and \textit{ft} indicate the number of pre-training steps and fine-tuning.}
    \label{tab:coco_karpathy}
\end{table*}

\begin{table*}[ht]
\small
    \centering
    \begin{tabularx}{0.7\linewidth}{l*{6}{>{\centering\arraybackslash}X}}
        \toprule
        & \multicolumn{3}{c}{\textbf{En}} & \multicolumn{3}{c}{\textbf{De}} \\
        \cmidrule(lr){2-4} \cmidrule(lr){5-7}
        & B@4 & C & CS & B@4 & C & MCS \\
        \midrule
        Gemma-3-12B & 17.8 & 50.8 & 84.0 & 17.3 & 55.5 & 87.0 \\
        Gemma-3-4B & 13.9 & 41.2 & 84.5 & 12.6 & 41.4 & 87.8 \\
        Pixtral-12B & 23.2 & 62.5 & 74.3 & 20.9 & 64.4 & 81.3 \\
        \midrule
        PaliGemma-3B & 33.2 & 88.9 & 77.2 & 17.3 & 57.6 & 79.3 \\ 
        \midrule
        Baseline-6B & 31.2 & 84.0 & 76.2 & 15.8 & 50.4 & 80.2 \\
        \midrule
        0.4B \textit{ft} (ours) & 28.6 & 76.6 & 75.6 & 28.2 & 77.4 & 80.9 \\
        0.4B \textit{100K ft} (ours) & 30.0 & 79.5 & 76.1 & 28.8 & 79.2 & 81.6 \\
        1.0B \textit{ft} (ours) & 30.2 & 85.1 & 77.1 & 28.6 & 82.6 & 82.5 \\
        3.5B \textit{ft} (ours) & 29.9 & 81.8 & 76.0 & 30.9 & 84.5 & 80.9 \\
        11.2B \textit{ft} (ours) & 30.0 & 83.1 & 76.2 & 33.2 & 90.7 & 80.8 \\
        11.2B \textit{30K ft} (ours) & 30.4 & 83.4 & 75.9 & 32.7 & 89.4 & 81.5 \\
        \bottomrule
    \end{tabularx}
    \caption{Image captioning results on the Multi30K~\cite{elliott2016multi30k} task 2 test 2016 split for English (En) and German (De) reporting the metrics BLEU-4 (B@4), CIDEr (C), CLIPScore (CS), and multilingual CLIPScore (MCS). \textit{100K}, \textit{30K}, and \textit{ft} indicate the number of pre-training steps and fine-tuning.}
    \label{tab:multi30k_ic}
\end{table*}

\begin{table*}[ht]
\small
    \centering
        \begin{tabularx}{0.7\linewidth}{l*{6}{>{\centering\arraybackslash}X}}
            \toprule
            & \textbf{En} & \textbf{De} & \textbf{Fr} & \textbf{Es} & \textbf{Ru} & \textbf{Zh} \\
            \midrule
            Gemma-3-12B & 34.0 & 39.6 & 61.3 & 52.7 & 43.9 & 28.4 \\
            Gemma-3-4B & 30.5 & 32.5 & 50.9 & 39.5 & 34.0 & 8.3 \\
            Pixtral-12B & 71.1 & 38.3 & 64.8 & 67.8 & 39.7 & 29.6 \\
            \midrule
            PaliGemma-3B & 79.1 & 37.7 & 64.7 & 67.1 & 35.1 & 27.1 \\ 
            \midrule
            Baseline-6B & 82.0 & 38.2 & 65.0 & 63.9 & 29.5 & 21.8 \\
            \midrule
            0.4B \textit{ft} (ours) & 76.9 & 38.6 & 55.4 & 54.9 & 26.0 & 23.7 \\
            0.4B \textit{100K ft} (ours) & 78.3 & 39.3 & 59.3 & 57.4 & 27.3 & 24.0 \\
            1.0B \textit{ft} (ours) & 78.0 & 41.2 & 64.6 & 59.6 & 30.3 & 24.2 \\
            3.5B \textit{ft} (ours) & 77.7 & 38.2 & 61.0 & 59.8 & 30.0 & 24.0 \\
            11.2B \textit{ft} (ours) & 79.3 & 39.4 & 64.1 & 61.6 & 34.3 & 25.3 \\
            11.2B \textit{30K ft} (ours) & 79.3 & 39.4 & 64.1 & 61.6 & 34.3 & 25.3 \\
            \bottomrule
        \end{tabularx}
    \caption{Multilingual image captioning results on XM3600~\cite{thapliyal2022xm3600}. \textit{100K}, \textit{30K}, and \textit{ft} indicate the number of pre-training steps and fine-tuning.}
    \label{tab:xm3600_ic}
\end{table*}

\FloatBarrier

\end{document}